\documentclass[%
 reprint,
groupedaddress,
 aps,
prstab,
longbibliography
]{revtex4-1}

\usepackage{graphicx, subcaption, amsmath, amssymb}
\usepackage{dcolumn}
\usepackage{bm}
\newcommand{\corr}[1]{\left\langle#1\right\rangle} 
\usepackage[T1]{fontenc}
\usepackage[utf8]{inputenc}
\usepackage[sort&compress]{natbib}
\usepackage[hidelinks]{hyperref}

\begin{document}


\title{Short sighted deep learning}

\author{Ellen de Mello Koch}
\email{ellen.demellokoch@wits.ac.za}

\author{Anita de Mello Koch}
\email{demellokochanita@gmail.com}

\author{Nicholas Kastanos}
\email{nicholaskastanos@gmail.com}

\author{Ling Cheng}
\email{Ling.Cheng@wits.ac.za}
\affiliation{%
School of Electrical and Information Engineering,
University of the Witwatersrand, Wits, 2050,
South Africa 
}

\date{\today}

\begin{abstract}
A theory explaining how deep learning works is yet to be developed.
Previous work suggests that deep learning performs a coarse graining, similar in spirit to the renormalization group (RG). 
This idea has been explored in the setting of a local (nearest neighbor interactions) Ising spin lattice. 
We extend the discussion to the setting of a long range spin lattice.
Markov Chain Monte Carlo (MCMC) simulations determine both the critical temperature and scaling dimensions of the system.
The model is used to train both a single RBM (restricted Boltzmann machine) network, as well as a stacked RBM network.
Following earlier Ising model studies, the trained weights of a single layer RBM network define a flow of lattice models.
In contrast to results for nearest neighbor Ising, the RBM flow for the long ranged model does not converge to the correct values for the spin and energy scaling dimension.
Further, correlation functions between visible and hidden nodes exhibit key differences between the stacked RBM and RG flows.
The stacked RBM flow appears to move towards low temperatures whereas the RG flow moves towards high temperature.
This again differs from results obtained for nearest neighbor Ising.

\end{abstract}

\maketitle

\section{Introduction}
\label{sec:intro}

In recent years machine learning has shown remarkable success in a wide range of problems. These algorithms can outperform humans at a number of tasks including image classification and complex strategy games \cite{krizhevsky2012imagenet,vinyals2019grandmaster,ranzato_factored_2010,salakhutdinov_restricted_2007,le2010deep,le2008representational}.
In many fields efforts to harness the power machine learning offers are being pursued. 
Specifically in physics, machine learning is being explored as a tool to uncover properties of systems that are difficult to study analytically \cite{carrasquilla2017machine,wang2016discovering,deng2017machine,broecker2017machine,aoki2016restricted,nomura2017restricted,morningstar2017deep,howard2018holographic}. 
However, there is at present no convincing explanation of how these networks learn and why they can achieve such remarkable feats \cite{paul_why_2014,zhang_understanding_2016,jordan2015machine,lin_why_2017}.

Recent studies have aimed to gain insight into how and why deep learning works, by employing the statistical mechanics of spin systems as simple toy models\cite{mehta_exact_2014,iso2018scale,koch2019deep,li2018neural, kim2018smallest, funai2018thermodynamics, koch2018mutual, beny_deep_2013}. 
The idea is that deep learning, which is a form of coarse graining, is related to the renormalization group (RG) of statistical mechanics and quantum field theory.
This is a natural idea since both problems are concerned with reducing many degrees of freedom to an effective description employing far fewer degrees of freedom.
Spin systems are an ideal setting in which to explore this proposal, thanks to their simplicity.
In addition to the simplicity of these models, their statistical mechanics is well understood making them a reasonable laboratory for deep learning studies.
Further, spin systems are also not far removed from more standard applications of deep learning including image analysis: the state of a spin system is composed of local interacting patches much in the same way that an image is made up of local collections of pixels which are highly correlated \cite{saremi_hierarchical_2013}.

Although a number of recent studies have explored the link between Kadanoff's variational renormalization group (RG) and deep learning \cite{mehta_exact_2014,iso2018scale,koch2019deep,li2018neural, kim2018smallest, funai2018thermodynamics, koch2018mutual, beny_deep_2013}, it is still not settled whether such a link exists.
Studies have so far focused on nearest neighbor Ising models. 
We add to this discussion by considering long range interacting spin systems which, as we will see, exhibit important differences from their short ranged cousins.

The paper \cite{iso2018scale} trained three (Restricted Boltzmann machines) RBM networks on Ising model data.
The first was trained on data generated at temperatures ranging from above to below the critical temperature, the second network on data above the critical temperature and the third at temperatures below the critical temperature.
The trained weights of each network were used to define a map from the visible spins back to the visible spins.
Iterating this map defines an RBM flow.
Results produced by the first network show interesting behavior suggesting that RBMs learn the critical temperature in the sense that the RBM flow terminates on the critical point of the Ising model.
This is in contrast to RG flows: the critical point of the Ising model is an unstable fixed point and fine tuning is required to construct an RG flow trajectory that terminates on this critical point. 

In a follow up paper, \cite{koch2019deep}, scaling dimensions are used to provide precise quantitative tests that probe whether or not the network reproduces the critical point dynamics.
The approach recognizes that the critical point of the Ising model is described by a conformal field theory (CFT), so that correlation functions of the spins in patterns generated by the RBM can be compared to known CFT predictions. 
The results of \cite{koch2019deep} show that the network correctly reproduces the smallest scaling dimension, which belongs to the Ising spin itself.
The network does not correctly reproduce the next smallest scaling dimension, which is related to the energy density.
This implies that although the RBM has correctly learned the largest scale correlations in the critical spin states, it fails on shorter length scales.

Our goal is to repeat the above investigations, for a two-dimensional long range spin lattice. 
Training a network on a system with a different range of interactions will allow us to probe how robust the claim is that RBMs learn the critical temperature of a given spin system and we can again probe what length scales are captured by training.
We use Markov Chain Monte Carlo methods (MCMC) to determine the critical temperature of the long range spin lattice and corresponding scaling dimensions, as well as to generate the RBM training data.
Our results suggest that the RBM flow does not converge to the critical temperature.
Further, stacking RBMs does not appear to generate a flow that matches the coarse features of RG flows.
Consequently, one lesson we learn is that it is dangerous to draw general conclusions from numerical experiments conducted on a single system.
Furthermore an interesting question which suggests itself (but which we will not manage to answer) is the question of what classes of models a given conclusion applies to.

The long range spin lattice we study is described in Section \ref{sec:longrange}. 
A quick overview of RG and RBMs is given in sections \ref{sec:RG} and \ref{sec:RBM}.
MCMC is discussed in Section \ref{sec:mcmc} along with the numerical approximations made for various properties of the long range spin lattice.
The results of our study are presented in Section \ref{sec:numerics}.

\subsection{\label{sec:longrange}Long range spin lattice model}

The long range spin lattice is a model of a magnet made up of binary spins. 
Considering the two-dimensional model we have a square lattice of sites with a spin at each site. 
Each spin, $s_{\bf{k}}$, can take values of $\pm 1$ and each site ${\bf k}$ is labelled by a two-dimensional vector of integer coordinates ${\bf k}=(x,y)$.
Spins on sites ${\bf i}$ and ${\bf j}$ interact with a strength of $J_{{\bf i}{\bf j}}$.
Each spin also interacts with an external magnetic field, with strength $\mu$.

The Hamiltonian of the long range spin lattice model is given by
\begin{equation}
H=-\sum_{\langle {\bf i},{\bf j}\rangle}J_{{\bf i}{\bf j}}s_{{\bf i}}s_{{\bf j}}-\mu\sum_{{{\bf j}}}s_{{\bf j}}.
\label{eq:h_ising}
\end{equation}
The interaction strength $J_{{\bf i}{\bf j}}$ is inversely proportional to a power of the distance between sites ${\bf i}$ and ${\bf j}$
\begin{equation}
J_{{\bf i}{\bf j}}=\frac{1}{r^{\alpha}}
\end{equation}
where $r=|{\bf i}-{\bf j}|$. We set $\alpha=3$ and $\mu=0$.

For the two-dimensional nearest neighbour Ising model, analytic results give precise values for the critical temperature and scaling dimensions.
When moving to models such as the long range spin lattice, no analytic results exist and we thus rely on MCMC methods to determine these properties before training the RBM \cite{geyer1992practical}.

The critical point of the long ranged spin system will again be described by a CFT.
The two point functions of primary operators of this CFT will have the usual power law fall off, something that can be explored numerically.
The two point function of primary operators at the fixed point of the spin lattice are of the form \cite{poland2019conformal}
\begin{equation}
  \corr{\phi(\vec x_1)\phi(\vec x_2)}={\frac{B}{|\vec x_1-\vec x_2|^{2\Delta}}}
  \label{eq:2point_corr}
\end{equation}
where $\Delta$ is the scaling dimension and $B$ is a constant.
To probe the largest scales of the patterns generated by the RBM we will focus on the two operators of the lowest dimension.
The first operator is the spin itself.
The two point correlator between spins $\corr{s_{\bf i}s_{\bf j}}$, with scaling dimension $\Delta_s$ takes the form
\begin{equation}
    \corr{s_{\bf i}s_{\bf j}}=\frac{B_{s}}{|{\bf i}-{\bf j}|^{2\Delta_s}}
    \label{eq:deltas}
\end{equation}
with $\Delta_s$ to be determined numerically.
The second operator of interest is the energy density where we study the correlator between $\corr{\epsilon_{\bf i} \epsilon_{\bf j}}$, with scaling dimension $\Delta_\epsilon$ 
\begin{equation}
    \corr{\epsilon_{\bf i}\epsilon_{\bf j}}=\frac{B_{\epsilon}}{|{\bf i}-{\bf j}|^{2\Delta_\epsilon}}
    \label{eq:deltaeps}
\end{equation}
with
\begin{equation}
    \epsilon_{{\bf i}}=s_{{\bf i}}\sum_{\langle {\bf i},{\bf j}\rangle}J_{{\bf i}{\bf j}}s_{\bf j}-\bar{\epsilon}_{\bf i},
\end{equation}
\begin{equation}
    \bar{\epsilon}_{\bf i}=\frac{1}{N_s}\sum_{ k=1}^{N_s}\epsilon_{\bf i}^{(k)}
\end{equation}
where $N_s$ is the number of sample configurations summed and $(k)$ denotes the $k$th sample.
We use MCMC sampling methods to determine the critical temperature of this model which is found to be $T_c \approx 7.7$ (details are discussed later in Section \ref{sec:mcmc}). 

\subsection{\label{sec:RG}RG}

RG defines a coarse graining transformation which is applied repeatedly to a microscopic description of a system to eventually arrive at a macroscopic description of the system \cite{wilson1974renormalization,efrati2014real}.
Each application of RG results in a rescaling where unimportant degrees of freedom are removed and important degrees of freedom remain.
The degrees of freedom or operators are characterized as either relevant, irrelevant or marginal.
At each step of applying RG, couplings between variables change and we obtain a new coarse grained Hamiltonian \cite{wilson1974renormalization}.
The only change that occurs when obtaining a new Hamiltonian at each step is the couplings update. 
The couplings that are relevant will become larger and the couplings that are irrelevant will get smaller as more steps of coarse graining are applied. 
The marginal terms in the Hamiltonian will remain unchanged.

\begin{figure}[ht!]
  \centering
  \includegraphics[width=0.2\textwidth]{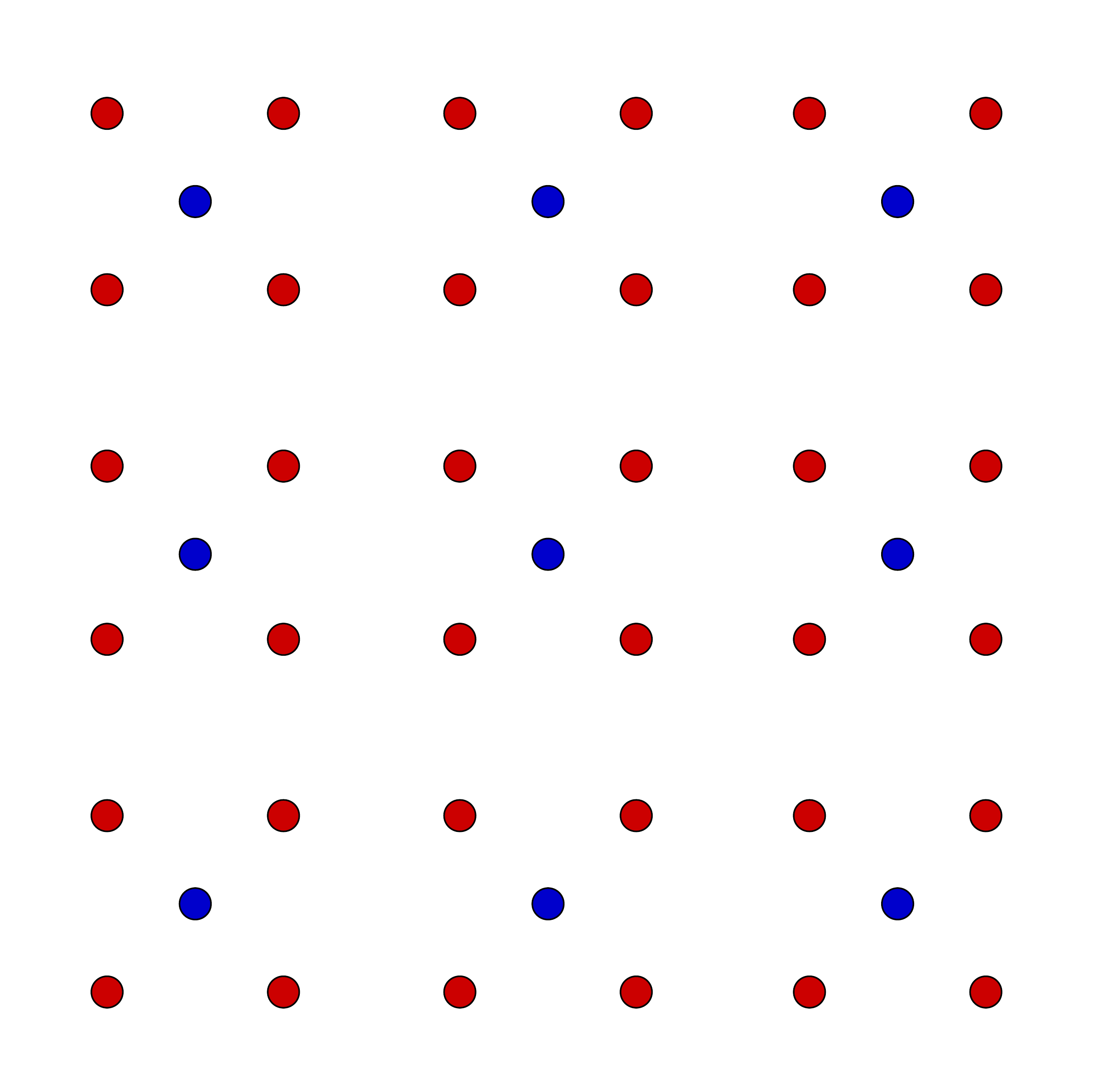}
  \caption{Block-spin averaging as introduced by Kadanoff. Groups of four red sites are averaged to obtain blue sites within their centre.}
  \label{fig:block_spin}
\end{figure}

In this paper we study a discrete spin lattice and so are interested in a discrete version of RG, defined by Kadanoff, which performs a block spin average \cite{kadanoff_variational_1976}.
Block spin averaging takes four spins as seen in Figure \ref{fig:block_spin} in red, and averages these spins to produce the blue spin found in the centre of each group of four red spins.
The average value of the red spins becomes the value of the new centre blue spin.
By performing this averaging we rescale lengths in the system by a factor of $1/2$.

The RG flow halts at a critical point where the system exhibits scale invariant properties.
This is because all couplings remain unchanged regardless of the length scale studied \cite{poland2019conformal}.
A basic observable of the fixed point is the spectrum of scaling dimensions.

\subsection{\label{sec:RBM}RBM}

Restricted Boltzmann machines are made up of two layers of nodes, an input layer called the visible layer and an output layer called the hidden layer. 
The number of nodes in the visible layer is dependant on the number of data points each training sample has. Selecting the number of nodes in the hidden layer is an iterative process, which tries to select architectures that give the best results. 
Nodes in both layers take values of $\pm 1$.
Every visible node is connected to every hidden node and connections between nodes in the same layer are forbidden.
The connection between a visible node $v_i$ and a hidden node $h_a$ has an associated weight $W_{ia}$.
Each visible node $v_i$ and hidden node $h_a$ has an associated bias $b_i^{(v)}$ and $b_a^{(h)}$ respectively.

The energy function of an RBM is given by
\begin{equation}
  E({\bf v},{\bf h})=-\sum_{i=1}^{N_v}\sum_{a=1}^{N_h}v_iW_{ia}h_a-\sum_{i=1}^{N_v}v_ib^{(v)}_i-\sum_{a=1}^{N_h}h_ab^{(h)}_{a}
\end{equation}
where $N_v$ is the number of visible nodes and $N_h$ is the number of hidden nodes.
This energy determines the probability of a visible and hidden vector occurring together, which is given by 
\begin{equation}
  P({\bf v},{\bf h})=\frac{e^{-E({\bf v},{\bf h})}}{Z}
\end{equation}
where 
\begin{equation}
  Z=\sum_{{\bf v},{\bf h}}e^{-E({\bf v},{\bf h})}
\end{equation}
is the partition function.
The goal of training is to match the distribution of the input data $q({\bf v})$ to the distribution generated by the 
RBM model $p({\bf v})$ \cite{hinton2012practical}, where
\begin{equation}
p({\bf v})=\frac{1}{Z}\sum_{\bf h}e^{-E({\bf v},{\bf h})}.
\end{equation}
To achieve this the weights and biases are updated using derivatives of the Kullback-Liebler divergence
\begin{equation}
  \begin{split}
    D_{KL}(q||p)&=\sum_{i=1}^{N_v} q(v_i)\left(\log\left(q({v_i})\right)-\log\left(p(v_i)\right)\right)\\
&=\sum_{i=1}^{N_v} q({v_i})\log\left(\frac{q({v_i})}{p(v_i)}\right).
  \end{split}
\end{equation}
Computation of the partition function $Z$ is practically impossible due to the enormous number of states summed over. 
To overcome this an approximate method, contrastive divergence, is used to train the network \cite{hinton2002training,hinton2012practical,sutskever2010convergence}.
Rather than summing over the entire space of vectors, we sum only over the vectors given in the training set \cite{carreira2005contrastive}.
During training the weights and biases are updated according to the following update rules
\begin{equation}
  \frac{\partial D_{KL}(q||p)}{\partial W_{ia}}=\langle v_ih_a \rangle_{data}-\langle v_ih_a \rangle_{model}
\label{eq:dklw}
\end{equation}
\begin{equation}
  \frac{\partial D_{KL}(q||p)}{\partial b_{i}^{(v)}}=\langle v_i \rangle_{data}-\langle v_i\rangle_{model}
\label{eq:dklbv}
\end{equation}
\begin{equation}
  \frac{\partial D_{KL}(q||p)}{\partial b_{a}^{(h)}}=\langle h_a \rangle_{data}-\langle h_a \rangle_{model}.
\label{eq:dklbh}
\end{equation}
We use values from the training data set to determine expectations given by $\langle \cdot \rangle_{data}$, and values from the model to determine expectations given by $\langle \cdot \rangle_{model}$.
To obtain the expectation of $\corr{v_i}_{data}$ we simply use the training data set, however we do not have a set of hidden vectors in the training set to determine $\corr{v_ih_a}_{data}$ or $\corr{h_a}_{data}$.
In order to find a set of data hidden vectors which we can use in the data expectations we sample hidden vectors using the training data set with a probability given by equation \eqref{eq:p_of_h}
\begin{equation}
  p({h_a}|{\bf v})=\tanh(\sum_{i}v_iW_{ia}+b^{(h)}_a).
  \label{eq:p_of_h}
\end{equation}  
We also do not have a set of visible vectors and hidden vectors for the model expectation values.
We use the data hidden vectors (generated using equation \eqref{eq:p_of_h} and the training data set) to generate model visible vectors with equation \eqref{eq:p_of_v}
\begin{equation}
  p({v_i}|{\bf h})=\tanh(\sum_{a}h_aW_{ia}+b^{(v)}_i).
  \label{eq:p_of_v}
\end{equation}  
Then using these model visible vectors, we can generate a set of model hidden vectors using equation \eqref{eq:p_of_h}.
This occurs at each step of training in order to calculate the updates given by equations \eqref{eq:dklw}, \eqref{eq:dklbv} and \eqref{eq:dklbh}.

%
%
Once training is complete, we can generate configurations, which we call an RBM flow, using the trained RBM \cite{iso2018scale,funai2018thermodynamics}.
Starting from a set of visible vectors ${\bf v}^{(1)}$ (could be the initial training data or MCMC data at a specific temperature), we generate a flow of visible and hidden vectors by sampling with the probabilities given in equations \eqref{eq:p_of_h} and \eqref{eq:p_of_v}, each time using the most recently generated set of vectors
\begin{equation}
  {\bf v}^{(1)}\rightarrow{\bf h}^{(1)} \rightarrow {\bf v}^{(2)}\rightarrow {\bf h}^{(2)} \rightarrow {\bf v}^{(3)}\rightarrow \cdots.
\end{equation}
In order to get ${\bf h}^{(1)}$ we would sample hidden nodes given the visible vectors ${\bf v}^{(1)}$. 
To get ${\bf v}^{(2)}$ we would sample visible nodes given the hidden vectors ${\bf h}^{(1)}$. 
We can iterate the process to generate an RBM flow. 
In Section \ref{sec:RBMflows} we discuss results derived from RBM flows.

\subsection{\label{sec:mcmc}Markov chain Monte Carlo}

Studying the long range spin lattice analytically is difficult.
We rely on MCMC numerical simulations to determine key statistical properties of the fixed point of the long range spin lattice. 
The fixed point theory is a conformal field theory, with a special class of operators, primary operators, that have two
point functions that have a power law fall off.
Probing these correlation functions provides detailed quantitative tests for the RBM output.
We are most interested in two primary operators: the basic spin and energy density operators given in 
equations \eqref{eq:deltas} and \eqref{eq:deltaeps}.

We begin with a randomly initialized lattice with each spin taking values of $\pm 1$ and a fixed temperature.
A single step of MCMC consists of the following:
\begin{enumerate}
  \item Randomly select a site $\bf i $.
  \item Determine the change in energy, $\Delta E$, that results from flipping $s_{\bf i}$.
  \item If $\Delta E\leq0$, accept the new configuration with probability 1.
  \item If $\Delta E>0$, accept the new configuration with probability $e^{-\Delta E/T}$.
\end{enumerate} 
We can generate a chain of lattice states by repeating the above process. 
Starting from a randomly initialized lattice means that a number of MCMC steps are required before we have a state that resembles states at the temperature we are concerned with. 
To ensure we sample valid states, allow a burn in period of sufficiently many MCMC steps before keeping samples generated.
The number of steps required for the burn in period is not known precisely but a good rule of thumb is to measure observables such as the magnetization or specific heat of the samples generated and ensure these values have stabilized before keeping samples generated.

\begin{figure}[ht!]
  \centering
  \begin{subfigure}[t]{0.35\textwidth}
    \includegraphics[width=\textwidth]{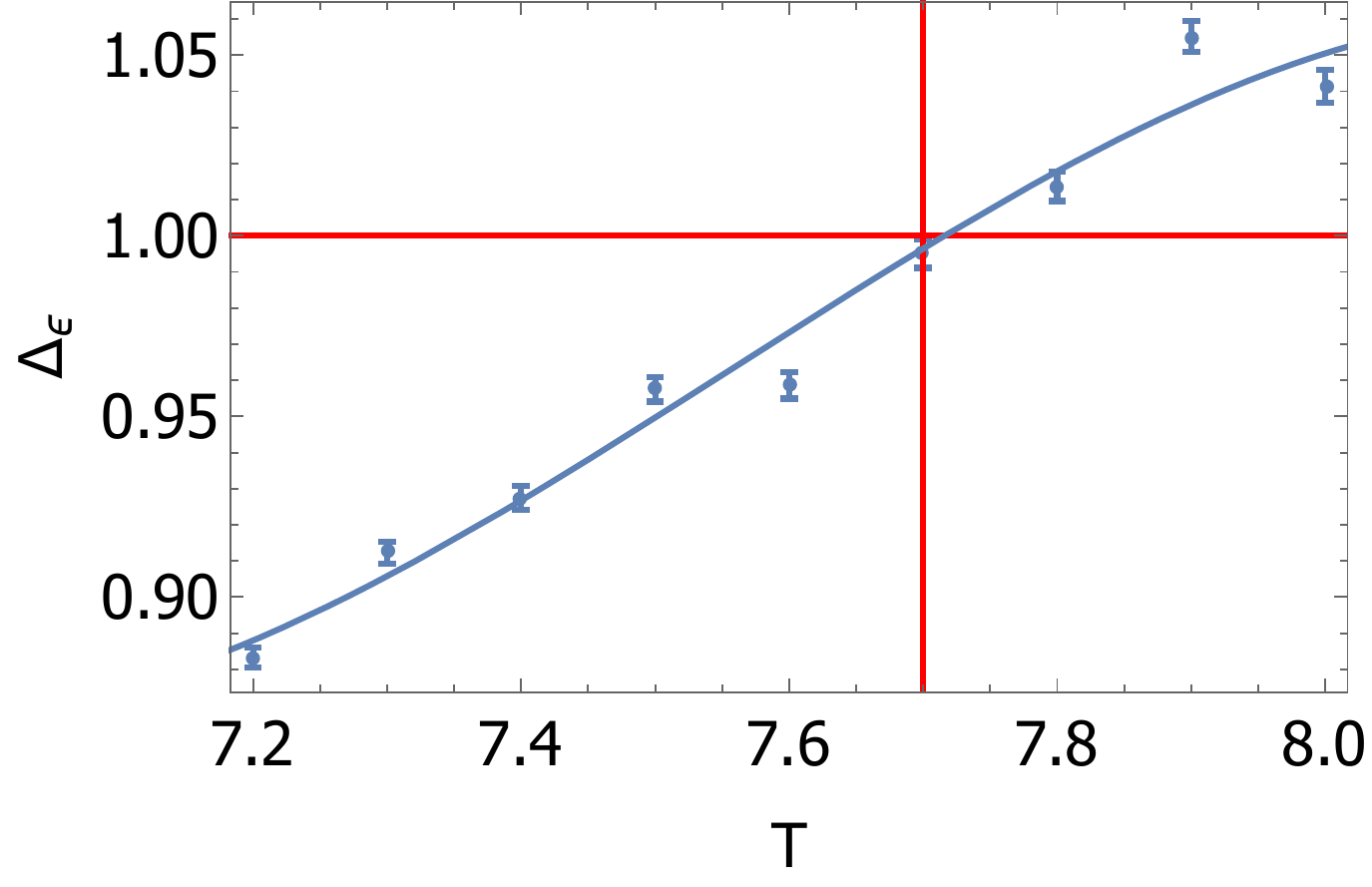}
    \caption{$\Delta_\epsilon$ versus temperature for the long range spin lattice with a lattice size of 10 by 10.}
    \label{fig:delta_e}
    \end{subfigure}
  \begin{subfigure}[t]{0.35\textwidth}
  \includegraphics[width=\textwidth]{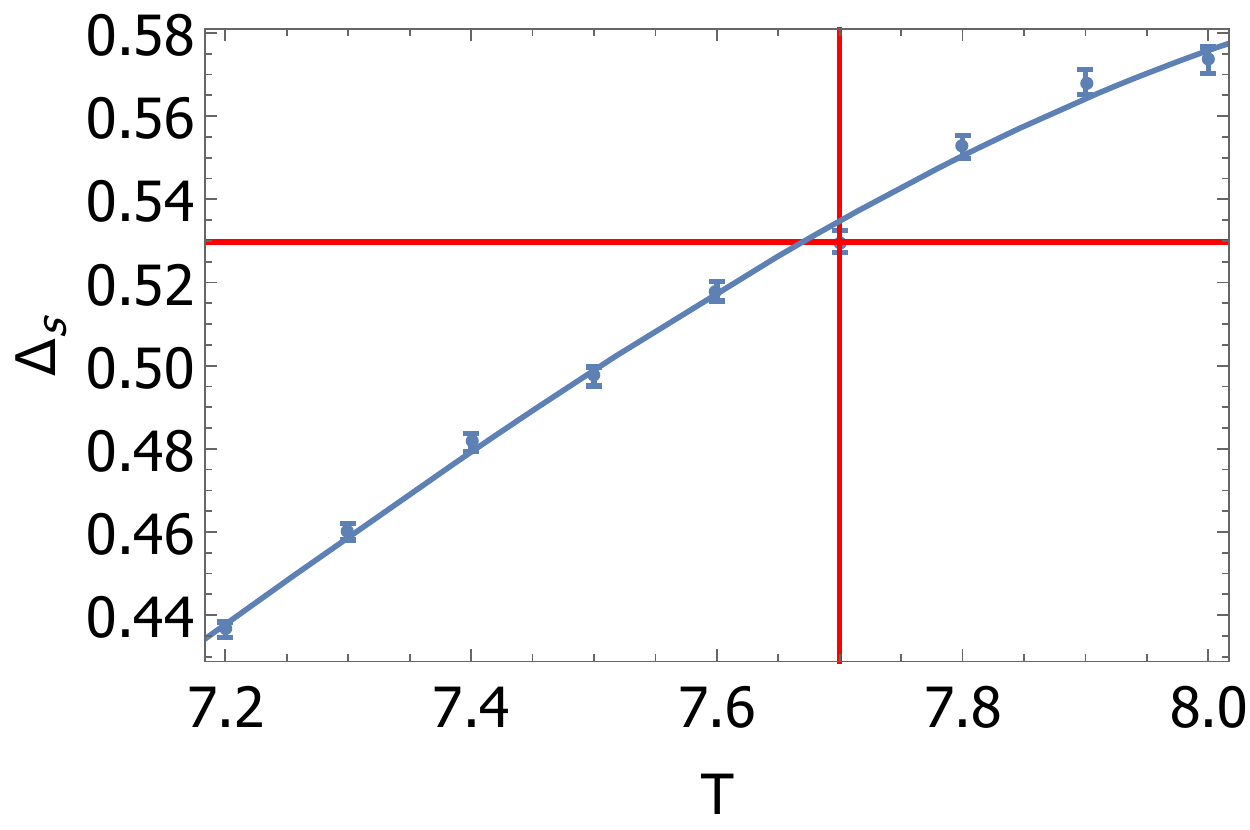}
  \caption{$\Delta_s$ versus temperature for the long range spin lattice with a lattice size of 10 by 10.}
  \label{fig:delta_S}
\end{subfigure}
  \begin{subfigure}[t]{0.38\textwidth}
  \includegraphics[width=\textwidth]{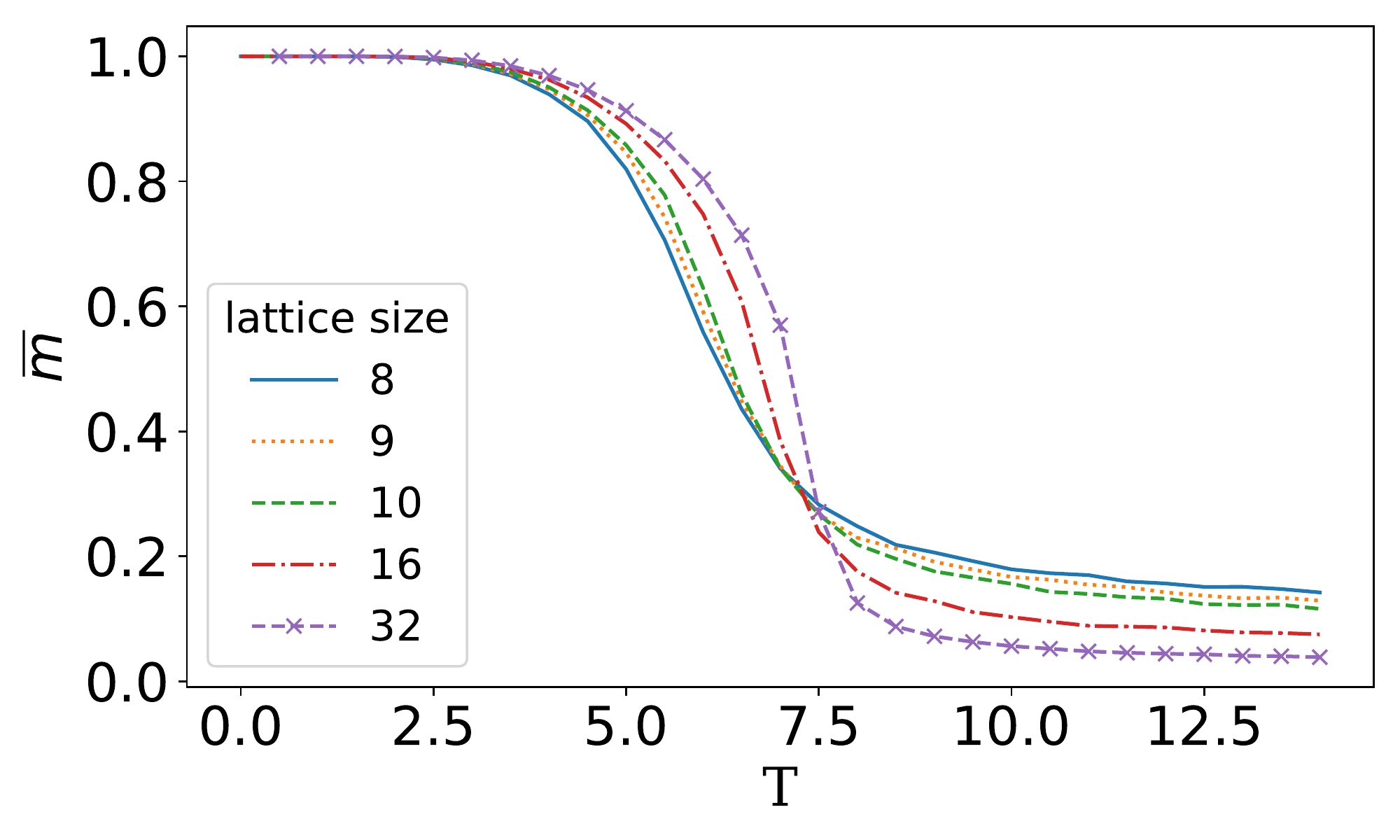}
  \caption{Magnetization versus temperature curve for various lattice sizes.}
  \label{fig:mag}
  \end{subfigure}
  \caption{Scaling dimensions of the spin lattice, with Hamiltonian in equation \eqref{eq:h_ising} with $\mu=0$ and 
$\alpha=3$, determined using MCMC. 
  $\Delta_\epsilon$ is 1 and $\Delta_s=0.53$ at $T=7.7$. 
  These values are at the intercept of the vertical and horizontal red lines in plots (a) and (b).}
  \label{}
\end{figure}

We generate 2000 samples at each temperature $T=0,0.1,0.2,\dots,14$ using a burn in period of 50000 MCMC steps. 
From the magnetization plot in Figure \ref{fig:mag} we see the critical temperature lies between $T=5$ and $T=8$.

We can give an independent determination of the critical temperature by studying correlation functions of  the primary operators given in equations \eqref{eq:deltas} and \eqref{eq:deltaeps}. 
The scaling dimension related to the energy, $\Delta_\epsilon$, must have a value of 1 at the critical point. 
From Figure \ref{fig:delta_e} we see that $\Delta_\epsilon =1$ when $T=7.7$. 
This value for the critical temperature lies in the correct range given by the magnetization plot.
Using 7.7 for $T_c$ we determine $\Delta_s$ from Figure \ref{fig:delta_S} which is shown by the vertical and horizontal lines as 0.53.

\section{\label{sec:numerics}Numerical results}

We perform two separate numerical experiments - one investigating a single RBM network 
and one investigating a simple "deep" network.
The first experiment studies the RBM flow, while the second explores the possibility that the layers in a stacked RBM 
network mimic steps in an RG flow.

The RBM flow is generated using a trained RBM.
The flow produces sets of visible vectors, one for each given initial vector, as described in Section \ref{sec:RBM}.
Our RBM flows for the two-dimensional long range spin lattice can be compared to existing results for the two-dimensional Ising spin lattice with nearest neighbour interactions.
As we will see, this comparison sheds light on the possible connection between RBM and RG flows.
Existing work shows that for the two-dimensional nearest neighbour Ising model, the RBM flows converge to the critical temperature of the system \cite{iso2018scale,funai2018thermodynamics}.
This is in contrast to RG flows because the Ising critical point is unstable and significant fine tuning is needed to ensure that a 
flow will terminate at the critical point.

Do RBM flows of the long range spin lattice flow to the critical point?
We have interrogated this question by comparing scaling dimensions for the spin and energy density operators, as well as temperatures, calculated using the configurations generated by the RBM flow.
By calculating these quantities for successive configurations in an RBM flow we determine if there is a fixed point of the flow and whether or not this fixed point is related to critical point of the long range spin lattice.
The results obtained from the RBM flows are reported in Section \ref{sec:RBMflows}.

The second experiment chooses the number of nodes in the stacked RBM networks in such a way that if there is a
correspondence to RG, the block spinning of each step combines spins in groups of 4.
This corresponds to scaling the input lattices by a factor of $1/2$.
To accomplish this, choose the number of hidden nodes to be $1/4$ of the number of visible nodes.
We study the flow generated by each RBM layer.
Each layer of the RBM network produces a set of hidden vectors that are input to the next RBM network in the stack.
Comparing each RBMs hidden layer configurations to steps in the coarse grained configurations produced by RG gives
a quantitative comparison between the two methods.

The stacked network of RBMs is trained in a greedy layer-wise manner \cite{bengio_greedy_2006}.
In greedy layer-wise training, each layer is trained independently and the hidden vectors produced by a trained layer are used as training data for the next layer.
Once all layers have been trained, we use the initial set of visible configurations  (generated from MCMC at $T_c$) and the hidden configurations from each layer to calculate correlations between visible and hidden nodes.
This $\corr{vh}$ correlator provides a diagnostic for comparing the coarse graining performed by RG versus that performed by
stacked RBM networks.

For the variational RG $\corr{{vh}}$ correlators for $v$ configurations we use the initial set of configurations generated by MCMC, while the $h$ configurations are generated by block spinning.
Results comparing RG to the stacked RBM are reported in Section \ref{sec:RGvsRBM}.

\subsection{\label{sec:RBMflows}RBM flows}

In this experiment we train a single RBM network and study the associated RBM flow. 
An RBM flow is a sequence of configurations generated by the trained network, starting from a set of initial visible configurations (see Section \ref{sec:RBM} for more details).
The RBM network we consider has $N_v=10\times 10=100$ visible neurons and $N_h=9\times 9=81$ hidden neurons. 
\begin{figure}[ht!]
  \centering
  \begin{subfigure}[t]{0.35\textwidth}
  \includegraphics[width=\textwidth]{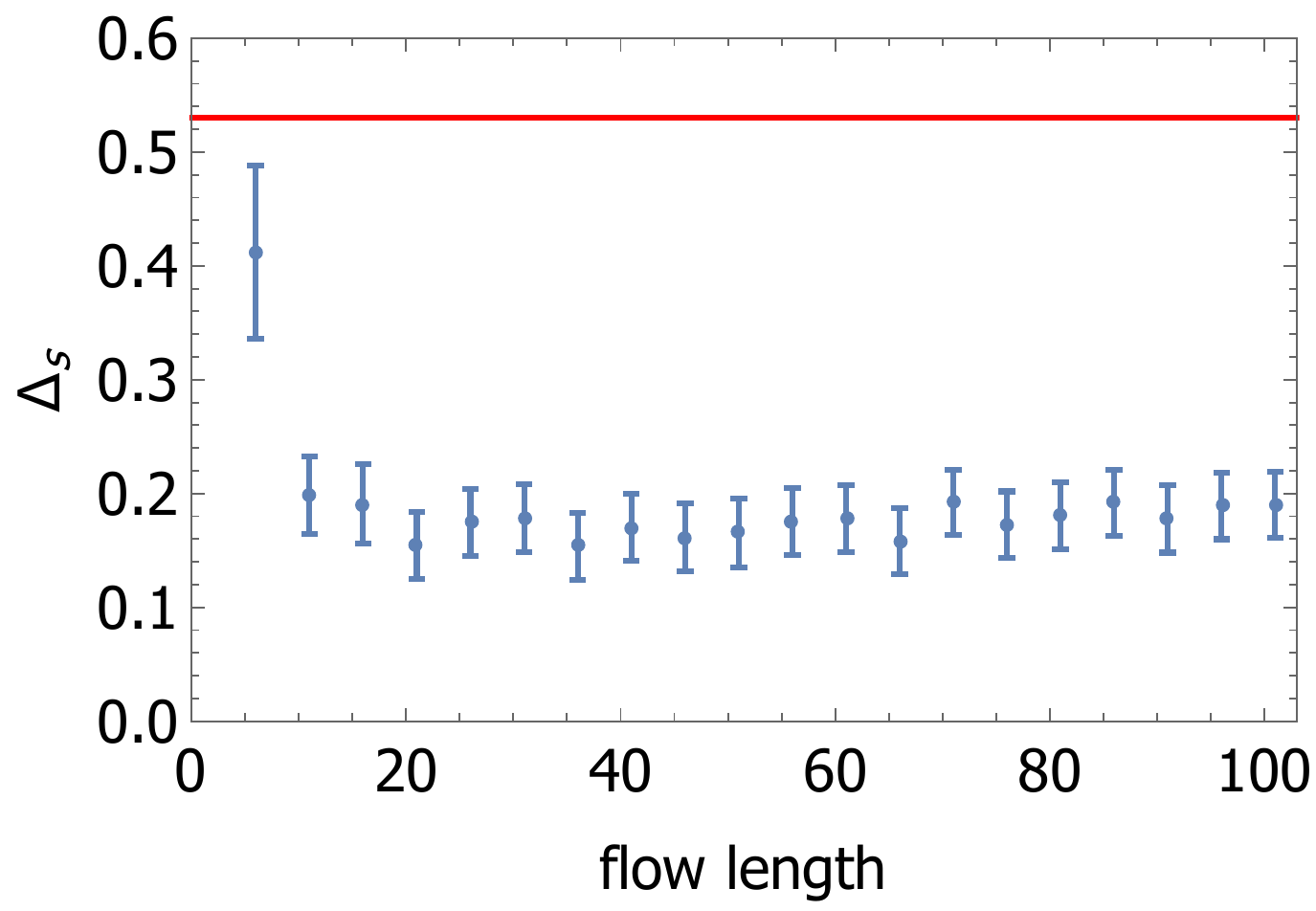}
  \caption{$\Delta_s$ versus flow length.}
  \label{fig:lowT_deltas}
  \end{subfigure}
  \begin{subfigure}[t]{0.35\textwidth}
    \includegraphics[width=\textwidth]{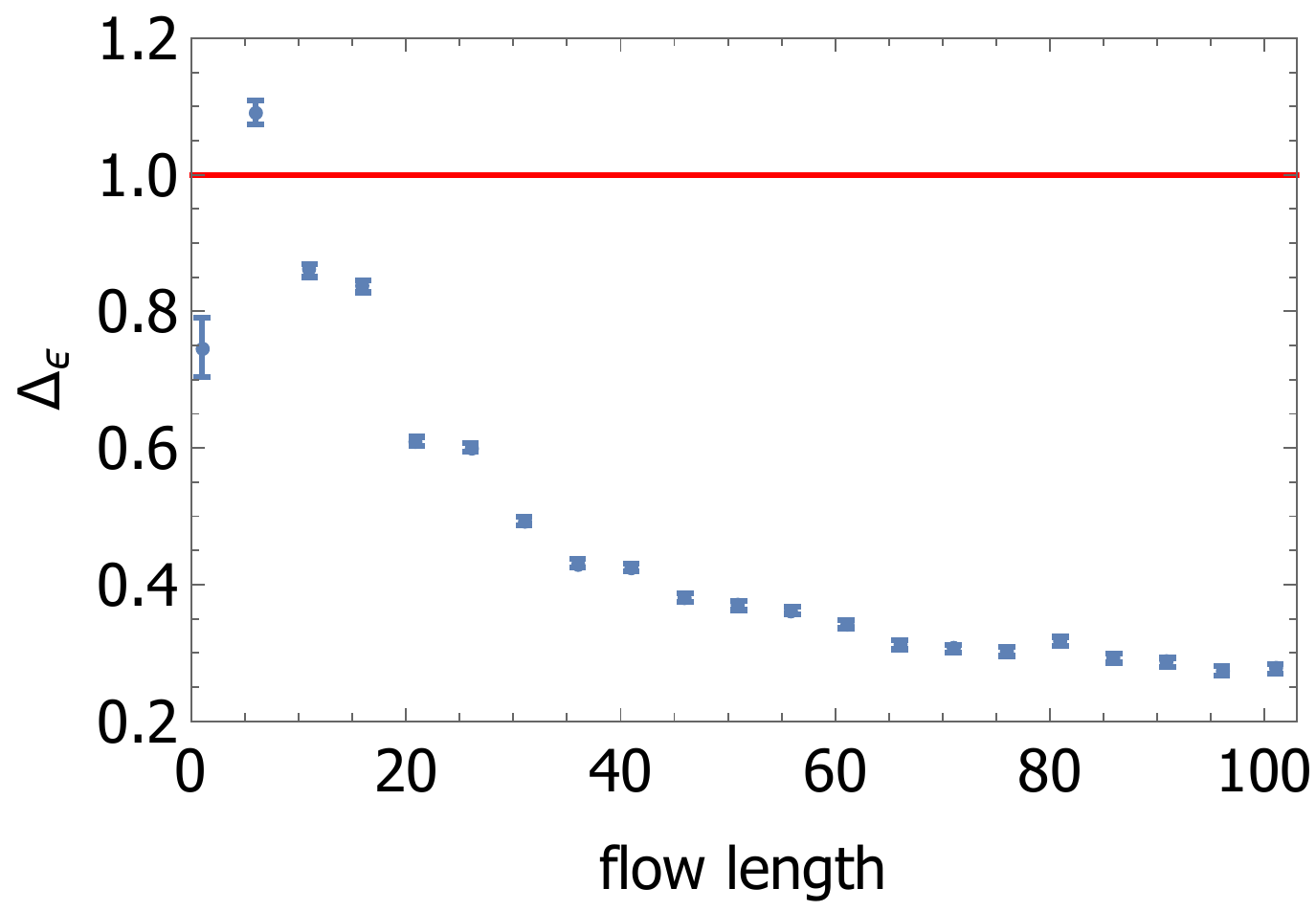}
    \caption{$\Delta_\epsilon$ versus flow length.}
    \label{fig:lowT_deltae}
    \end{subfigure}
    \begin{subfigure}[t]{0.35\textwidth}
      \includegraphics[width=\textwidth]{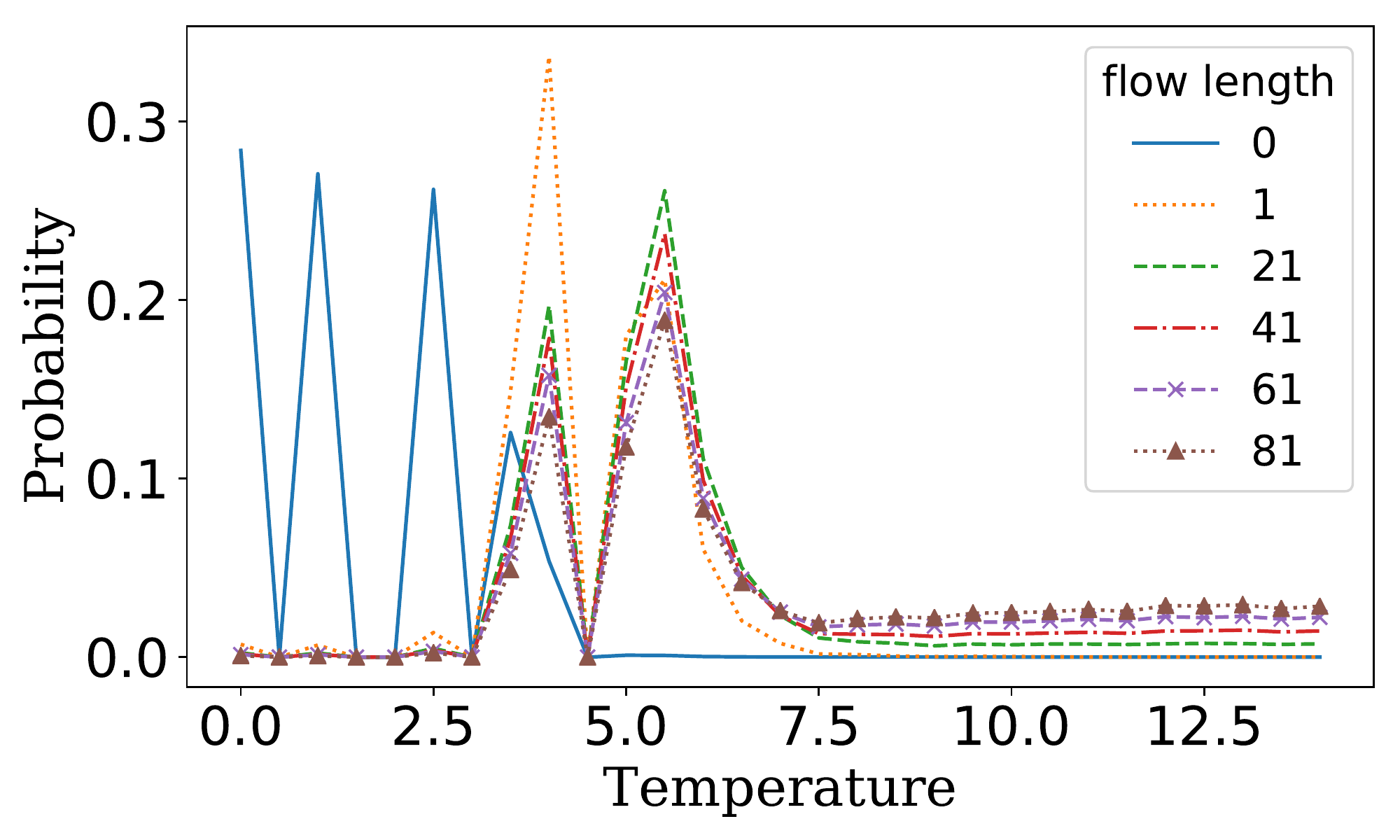}
      \caption{Average probability of temperature versus temperature of flows of various lengths. Flow tends to a temperature of 6.}
      \label{fig:lowT_temp}
      \end{subfigure}
  \caption{RBM flows starting from low temperature ($T=0$) configurations.
  The RBM has $N_v=100$ visible nodes and $N_h=81$ hidden nodes.
  The training set is comprised of 2000 configurations at each temperatures $T=0,0.5,\dots,14$ giving 30000 configurations in
total.}
  \label{fig:low_T}
\end{figure}

\begin{figure}[ht!]
  \centering
  \begin{subfigure}[t]{0.35\textwidth}
  \includegraphics[width=\textwidth]{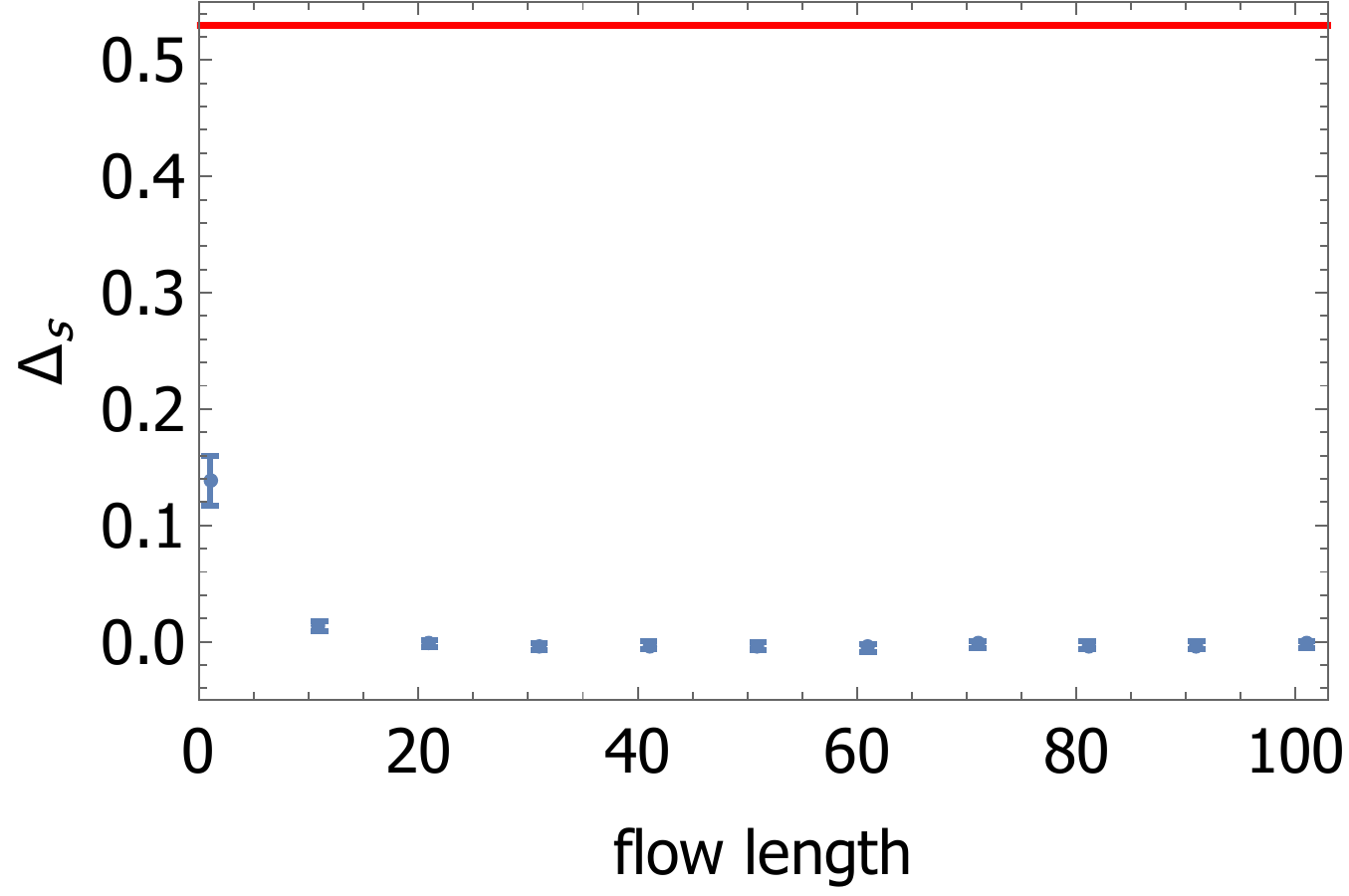}
  \caption{$\Delta_s$ versus flow length.}
  \label{fig:Tc_deltas}
  \end{subfigure}
  \begin{subfigure}[t]{0.35\textwidth}
    \includegraphics[width=\textwidth]{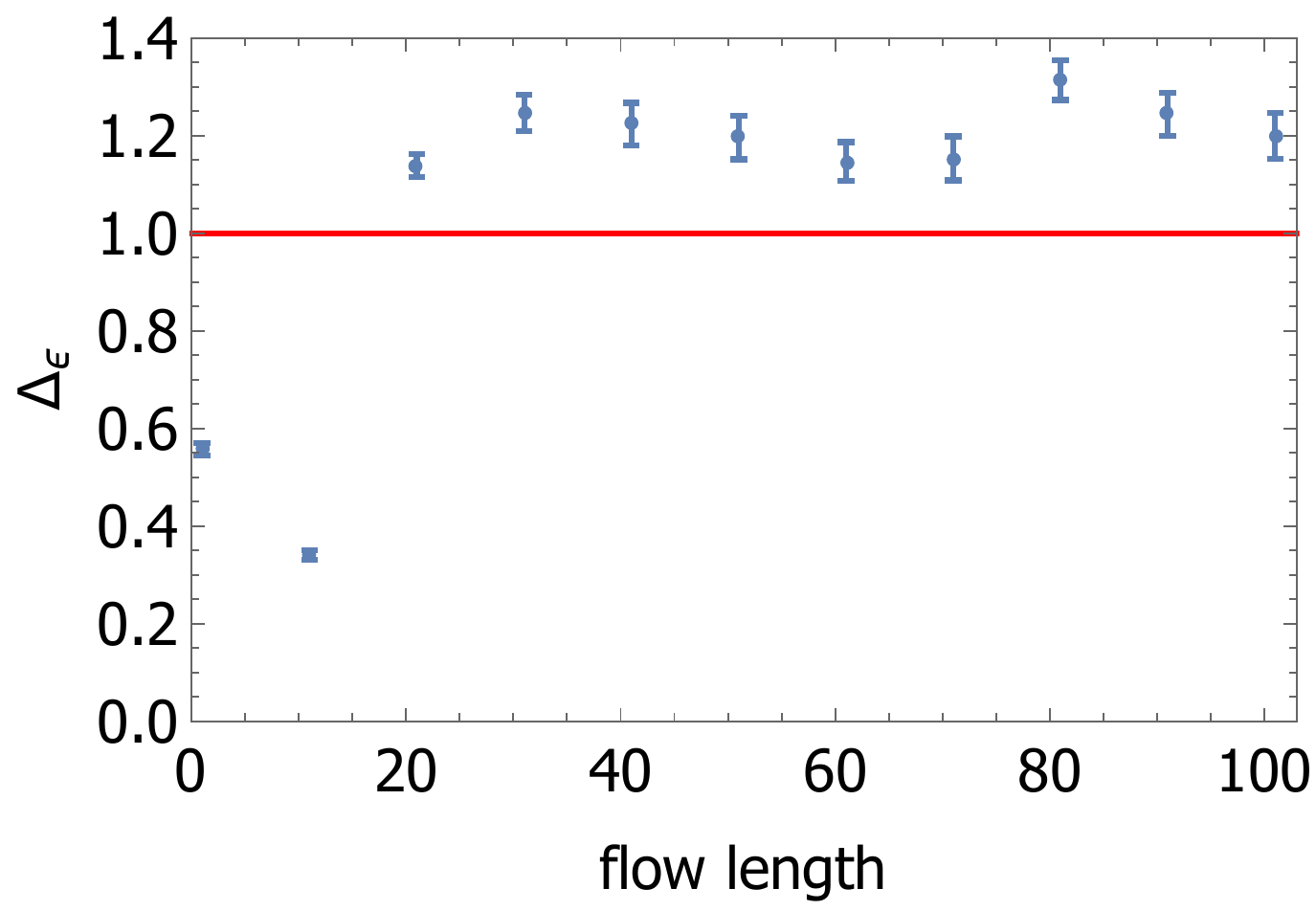}
    \caption{$\Delta_\epsilon$ versus flow length.}
    \label{fig:Tc_deltae}
    \end{subfigure}
    \begin{subfigure}[t]{0.35\textwidth}
      \includegraphics[width=\textwidth]{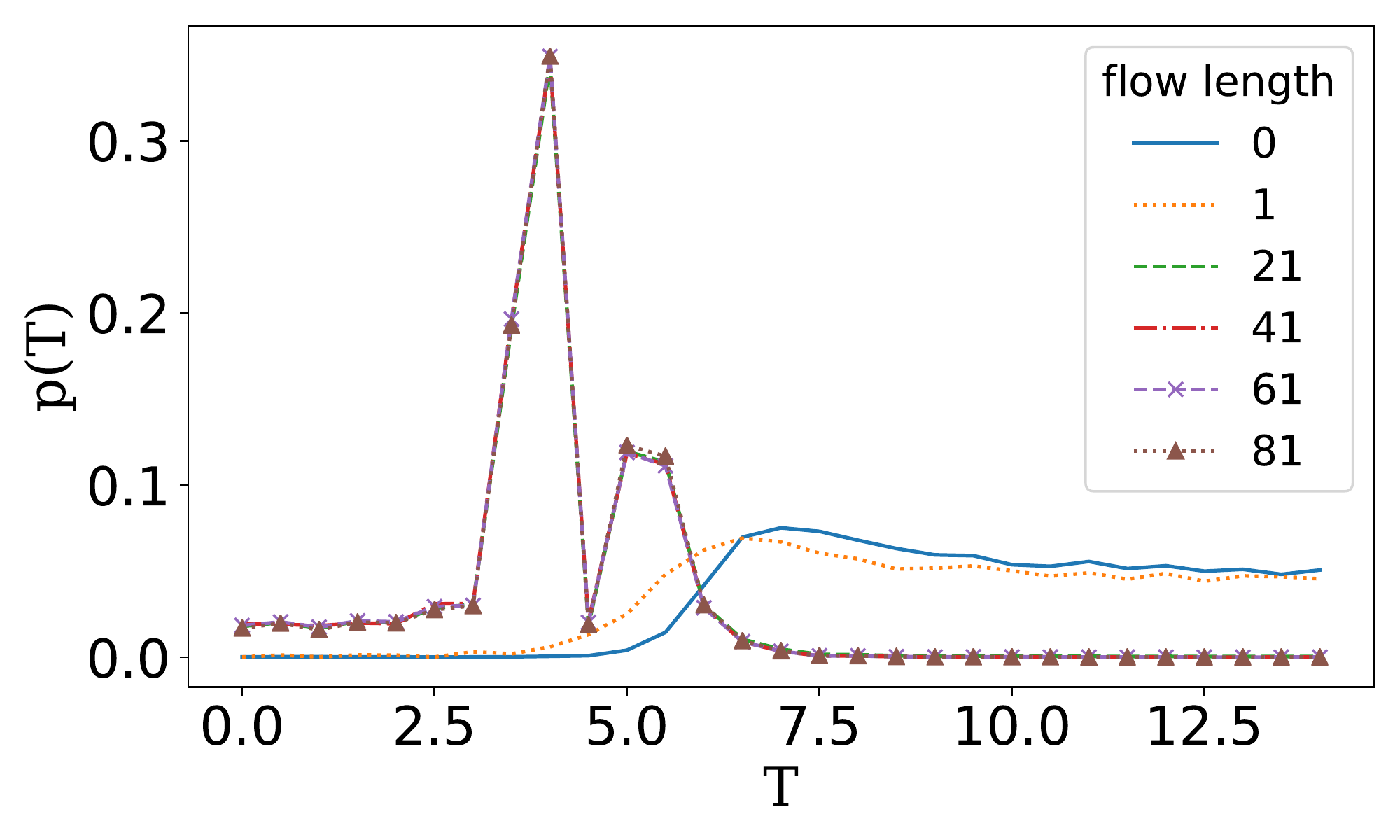}
      \caption{Average probability of temperature versus temperature of flows of various lengths.}
      \label{fig:Tc_temp}
      \end{subfigure}
  \caption{RBM flows starting from configurations near the critical temperature ($T_c\approx T=7.7$).
  The RBM has $N_v=100$ visible nodes and $N_h=81$ hidden nodes.
  Training uses 2000 configurations at temperatures $T=0,0.5,\dots,14$ giving 30000 configurations in total.}
  \label{fig:Tc}
\end{figure}

The data used to train the network is generated using MCMC using the long range spin lattice Hamiltonian.
2000 configurations are generated at each temperature $T=0,0.5, \dots ,14$. 
The network is trained using 30000 training steps, a learning rate of $10^{-3}$ and batches of 1000 samples.
Once the network has been trained, we generate flows starting from (i) low temperature $T=0$, (ii) the critical temperature $T_c=7.7$ and (iii) high temperature $T=14$.
The configurations generated for different flow lengths determine the scaling dimensions of the spin, $\Delta_s$, and energy density, $\Delta_\epsilon$, operators. 
In addition, by employing a supervised neural network, we also measure the average temperature of flows at different lengths.
This allows us to determine both how the dimensions and the temperature evolve with the RBM flow. 
The supervised network is trained on the same data as is used to train the RBM. 
Note however that for the supervised case this data contains temperature labels.

\begin{figure}[ht!]
  \centering
  \begin{subfigure}[t]{0.35\textwidth}
  \includegraphics[width=\textwidth]{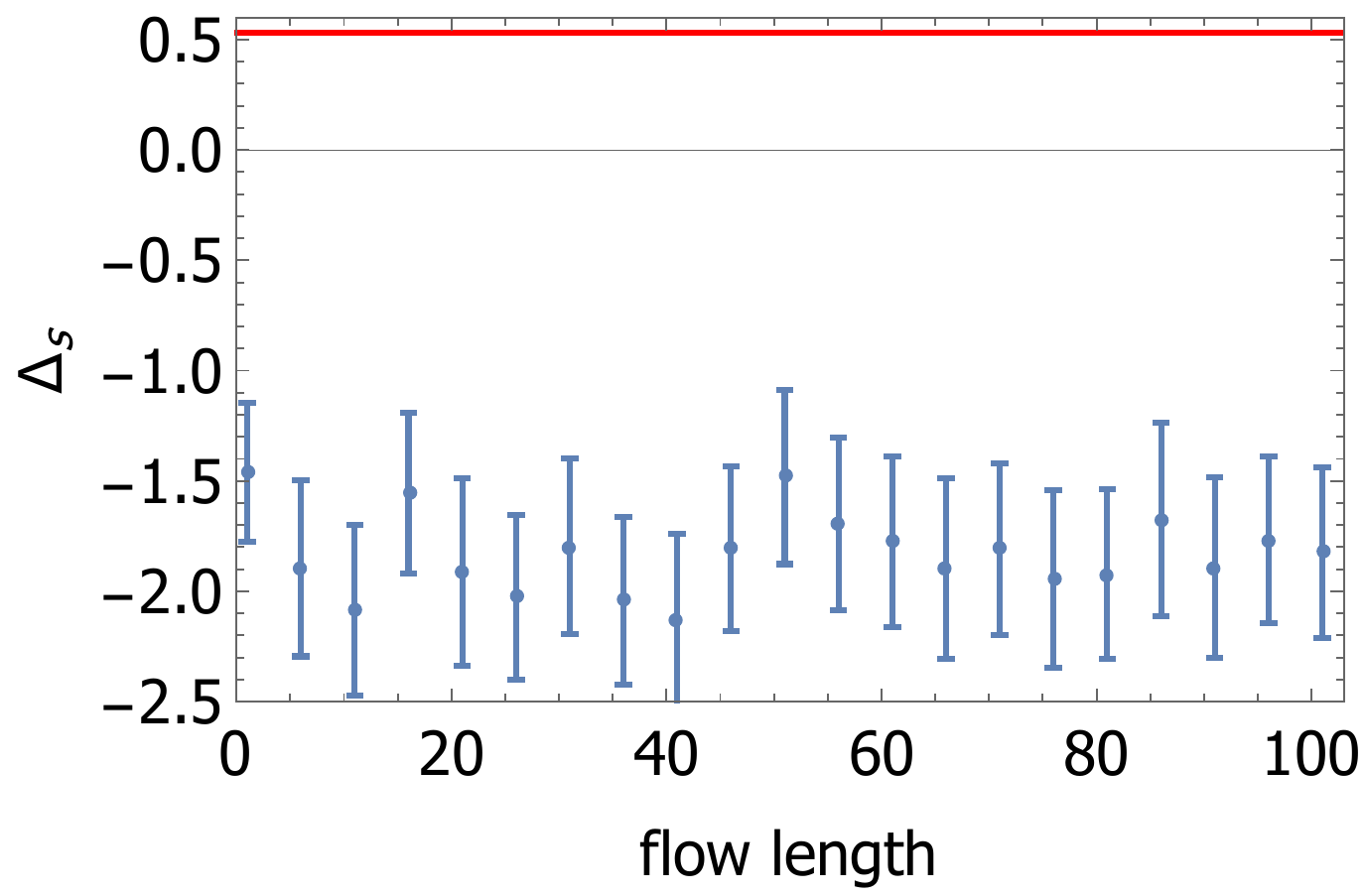}
  \caption{$\Delta_s$ versus flow length.}
  \label{fig:highT_deltas}
  \end{subfigure}
  \begin{subfigure}[t]{0.35\textwidth}
    \includegraphics[width=\textwidth]{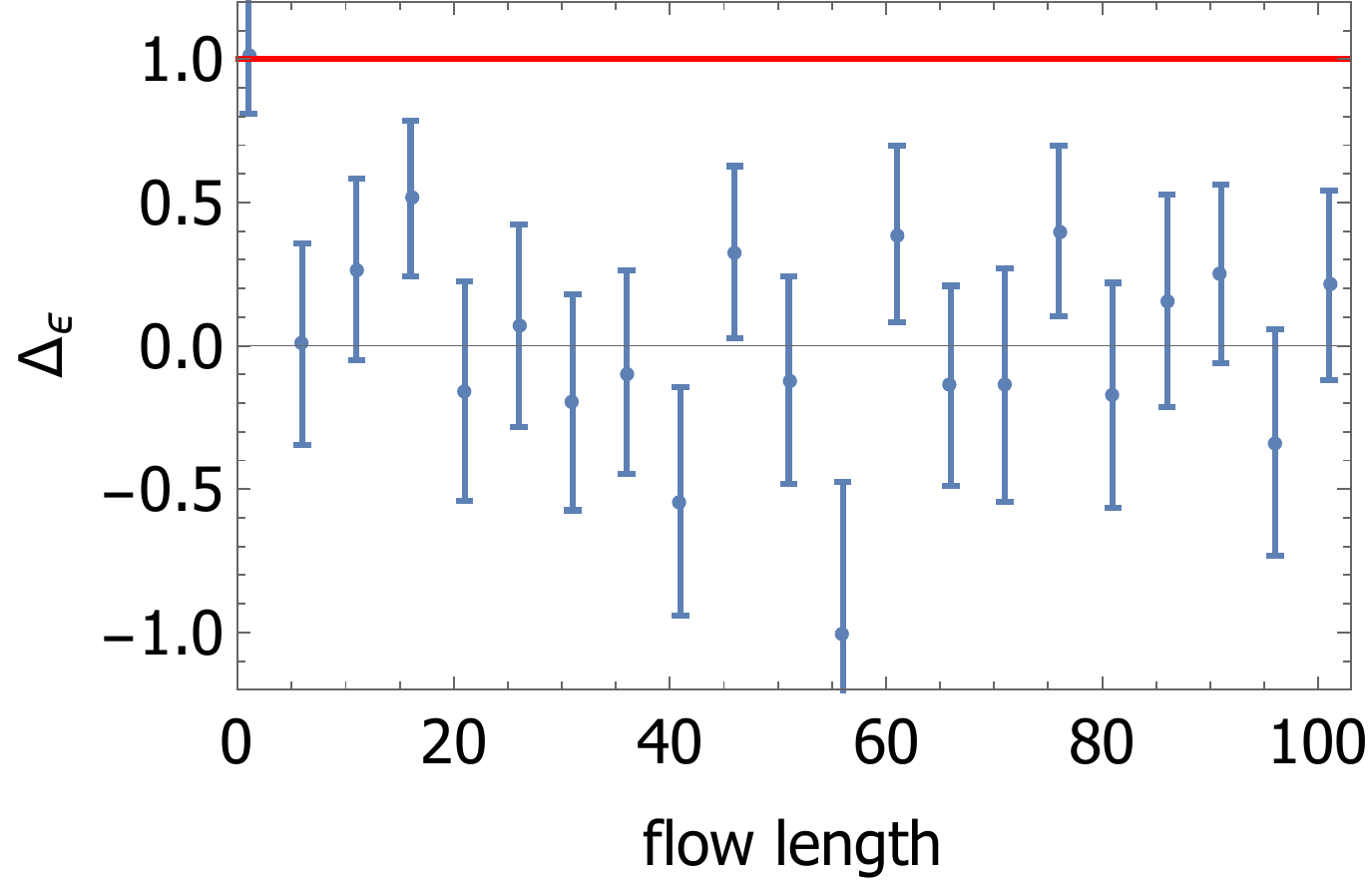}
    \caption{$\Delta_\epsilon$ versus flow length.}
    \label{fig:highT_deltae}
    \end{subfigure}
    \begin{subfigure}[t]{0.35\textwidth}
      \includegraphics[width=\textwidth]{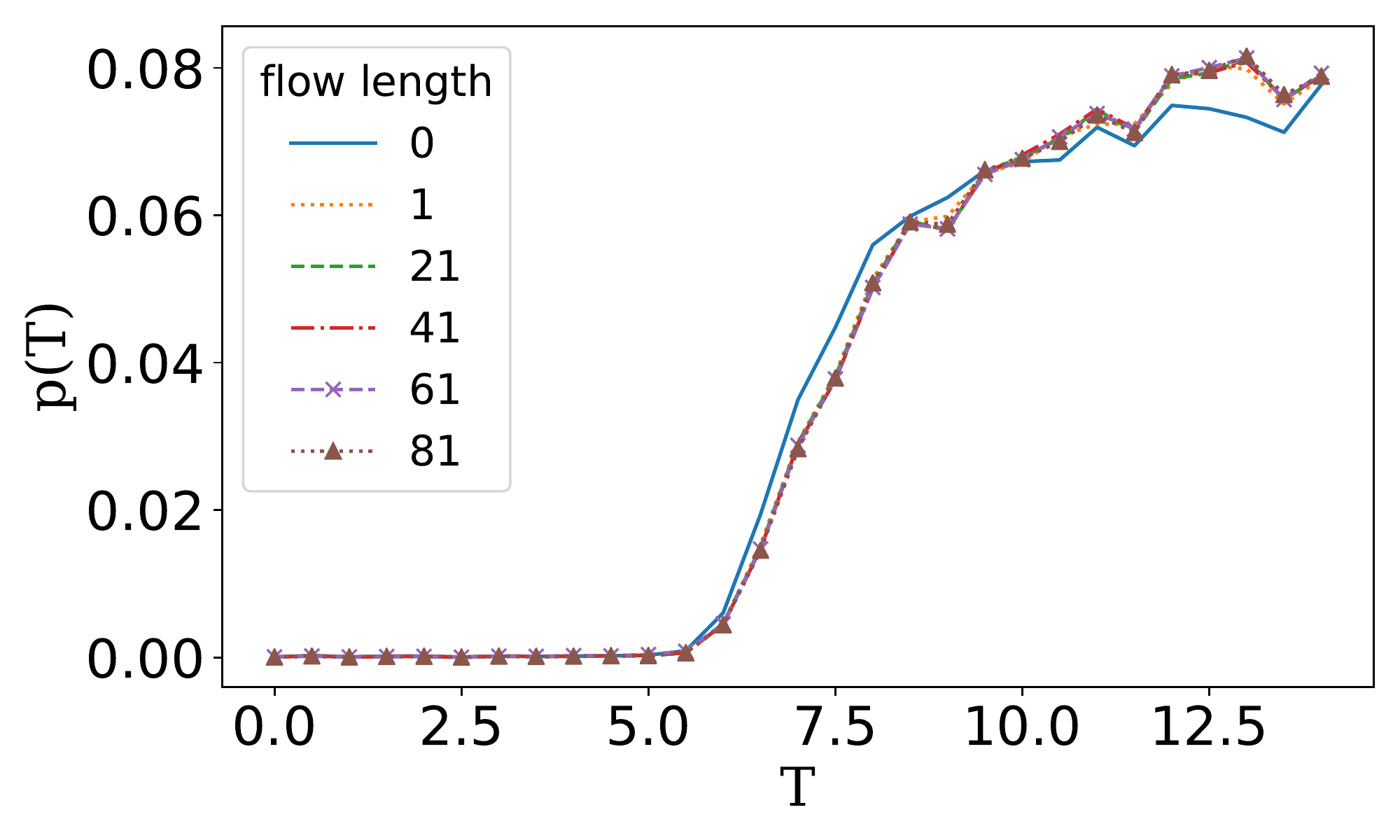}
      \caption{Average probability of temperature versus temperature of flows of various lengths.}
      \label{fig:highT_temp}
      \end{subfigure}
  \caption{RBM flows starting from configurations at a high temperature ($T=14$).
  The RBM network has $N_v=10\times10=100$ visible nodes and $N_h=9\times9=81$ hidden nodes.
  The training set is made up of configurations at temperatures of $T=0,0.5,\dots,14$ with 2000 configurations at each temperature (in total 30000 configurations).}
  \label{fig:high_T}
\end{figure}

Results for flows starting from (i) low temperature configurations are given in Figure \ref{fig:low_T}, (ii) the critical temperature in Figure \ref{fig:Tc} and (iii) high temperature in Figure \ref{fig:high_T}. 
Figure \ref{fig:lowT_deltas} converges after a flow length of $\pm 20$ to a value for $\Delta_s$ of approximately $0.175$ which underestimates the value determined from MCMC of $0.53$ shown by the red horizontal line.
This implies that the patterns generated by the RBM are more correlated on large length scales than they should be. 
In contrast to this, results obtained in \cite{koch2019deep} show that flows generated by an RBM trained on two-dimensional Ising model configurations, starting from low temperature configurations, converge to the correct value for $\Delta_s$ \cite{koch2019deep}.
There is not improvement when these RBM configurations are used to estimate the scaling dimension of the energy density: Figure \ref{fig:lowT_deltae} shows that the flows generate a value of $\Delta_\epsilon=0.25$, which is again an under estimate of the correct value $\Delta_\epsilon=1$.
The RBM has consistently underestimated the scaling dimensions, implying that distant spins are more correlated 
than they should be.
Intuitively, more correlation suggests that we are below the critical temperature and we indeed find that the flows converge to a temperature of approximately 6 which is below $T_c$. 
A configuration below the critical temperature, has more correlation between distant spins than those at the critical temperature.

Clearly the RBM has not correctly determined the critical point of the long range spin lattice, which is in tension with conclusions reached when studying the nearest neighbor Ising spin lattice: in that case flows converge to the critical temperature  \cite{iso2018scale,funai2018thermodynamics}.

Starting the flow from configurations at the critical temperature gives the results shown in Figure \ref{fig:Tc_deltas}.
The flows converge to a value close to 0 for $\Delta_s$, well below the critical point value. 
For $\Delta_\epsilon$ the flow converges to a value near 1.2, as shown in Figure \ref{fig:Tc_deltae}.
This is an over estimate of the critical point value $\Delta_\epsilon=1$. 
Figure \ref{fig:Tc_temp} shows that the flow converges to a temperature of $3.75$, again lower than the critical temperature.
These results suggest that the configurations generated by the RBM flow have not managed to capture the physics of the long
ranged spin lattice. 

For the RBM flow starting from high temperature, results in Figure \ref{fig:highT_deltas} suggest that $\Delta_s<0$, which is completely unphysical: this corresponds to a situation when the correlation between two spins grows with the distance between the spins.
The results in Figure \ref{fig:highT_deltae} suggest that the $\Delta_\epsilon$ value does not converge.
In the flow starting from high temperature configurations, the temperature appears to remain high as the RBM flow length increases.

\subsection{\label{sec:RGvsRBM}Stacked RBM and RG comparison}

In this section we study the flow generated by successive stacked layers of a 3-layer deep RBM network. 
The first layer has $N_v=64\times64 = 4096$ visible and $32\times32=1024$ hidden nodes, the second  $32\times32=1024$ visible and $16\times16=256$ hidden nodes and the third $16\times16=256$ visible and $8\times8=64$ hidden nodes.
The networks are trained in a greedy layer wise manner, i.e. each layer is trained separately and in order.

The first network is trained on $30000$ configurations at a temperature of $T=7.7\approx T_c$ with lattice size $64\times64$ \cite{bengio_greedy_2006}.
Once this network is trained, hidden configurations are generated by feeding training data to the first network.
The hidden configurations generated by the first RBM are the training data for the second RBM.
Once trained the second RBM generates training data for the third and final layer.

To test if the stacked RBM flow resembles an RG flow, we compare to three steps of an RG flow.
The deep network we have studied eliminates a quarter of the spins in each layer, which we mirror by block spinning groups of four spins.
Consequently, the first step of RG will coarse grain lattices with $N_v=64\times64=4096$ spins to lattices with $32\times32=1024$ spins, the second step then coarse grains to lattices with $16\times16=256$ spins and the third to lattices with $8\times8=64$ spins.

To develop a quantitative comparison of the flows of the stacked RBM and those of block spin RG, we calculate $\corr{vh}$ correlators between the initial spins populating the $64\times64=4096$ lattice and the spins populating each subsequent coarse grained lattice.
The result is a matrix of values, $\corr{v_ih_a}$, indexed by a label $i$ for a site of the original lattice and a label $a$ for a site of a coarse grained lattice. 
Selecting a specific column, for example $\corr{v_ih_1}$, gives the correlation of a specific coarse grained spin with all of the
original spins.
Rearranging the resulting $\corr{v_ih_1}$ vector into a $64\times64$ matrix, produces a visual pattern of the correlation
between the coarse grained spin and the original spins.
In this way our results are used to produce a picture of how the coarse graining is achieved. 

\begin{figure}[h!]
  \centering
  \begin{subfigure}{0.11\textwidth}
    \includegraphics[width=\textwidth]{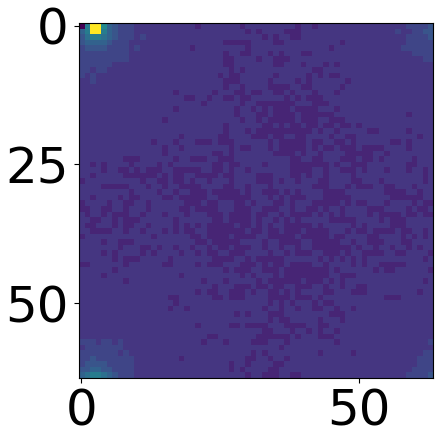}
    \caption{}
    \label{}
  \end{subfigure}
  \begin{subfigure}{0.11\textwidth}
    \includegraphics[width=\textwidth]{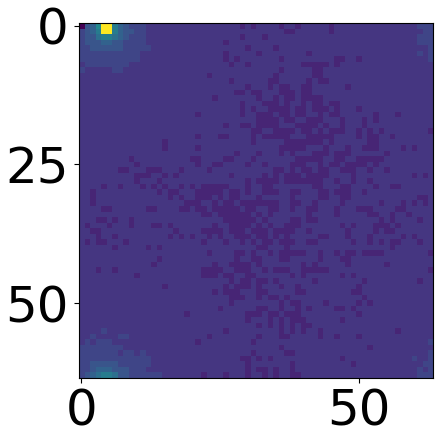}
    \caption{}
    \label{}
  \end{subfigure}
  \begin{subfigure}{0.11\textwidth}
    \includegraphics[width=\textwidth]{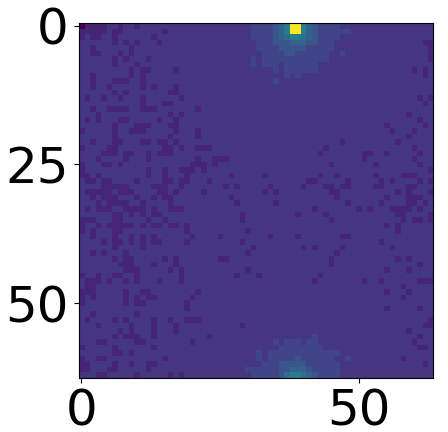}
    \caption{}
    \label{}
  \end{subfigure}
    \begin{subfigure}{0.11\textwidth}
    \includegraphics[width=\textwidth]{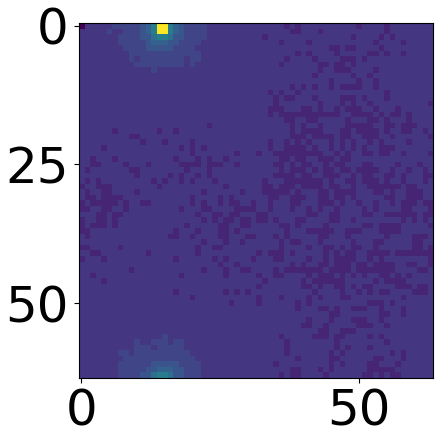}
    \caption{}
    \label{}
  \end{subfigure}

  \begin{subfigure}{0.11\textwidth}
    \includegraphics[width=\textwidth]{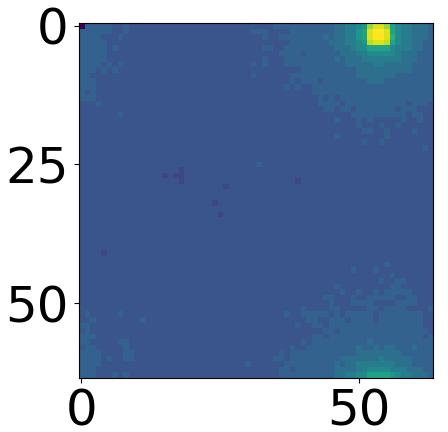}
    \caption{}
    \label{}
  \end{subfigure}
  \begin{subfigure}{0.11\textwidth}
    \includegraphics[width=\textwidth]{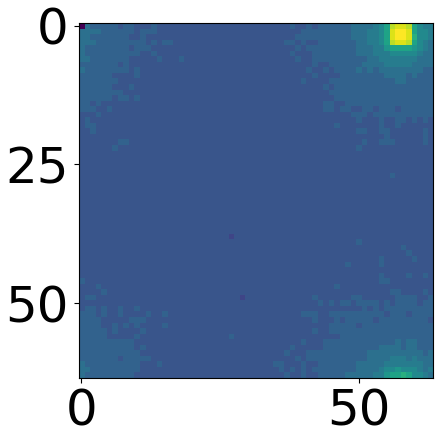}
    \caption{}
    \label{}
  \end{subfigure}
  \begin{subfigure}{0.11\textwidth}
    \includegraphics[width=\textwidth]{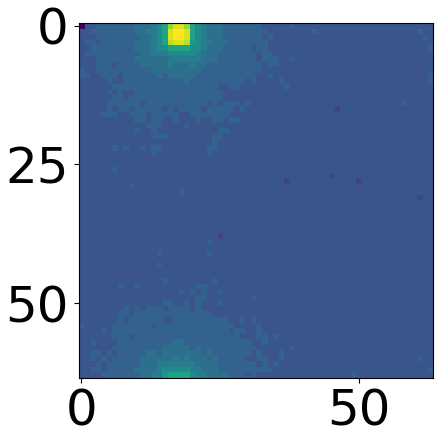}\caption{}
    \label{}
  \end{subfigure}
  \begin{subfigure}{0.11\textwidth}
    \includegraphics[width=\textwidth]{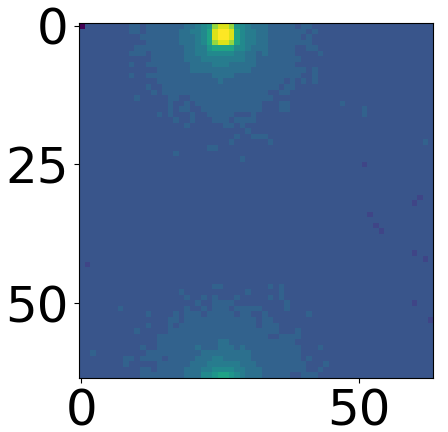}
    \caption{}
    \label{}
  \end{subfigure}

  \begin{subfigure}{0.11\textwidth}
    \includegraphics[width=\textwidth]{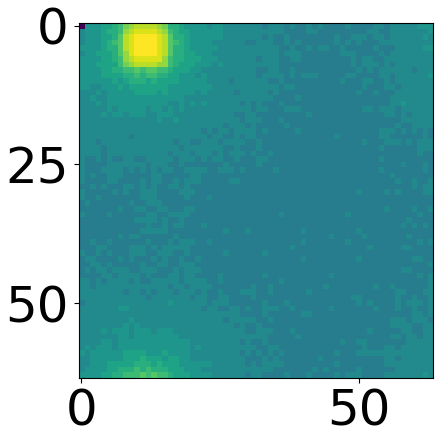}
    \caption{}
    \label{}
  \end{subfigure}
  \begin{subfigure}{0.11\textwidth}
    \includegraphics[width=\textwidth]{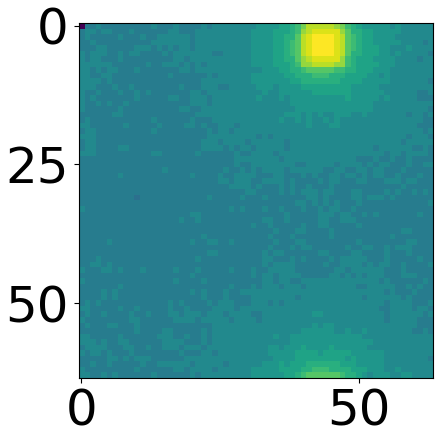}
    \caption{}
    \label{}
  \end{subfigure}
  \begin{subfigure}{0.11\textwidth}
    \includegraphics[width=\textwidth]{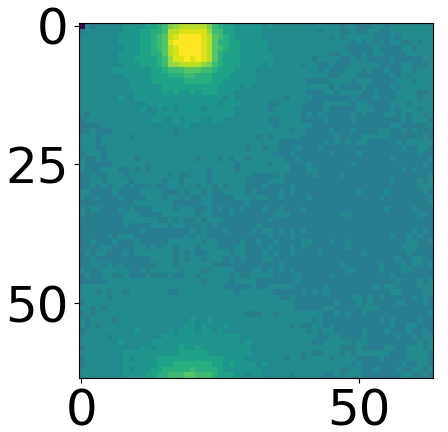}
    \caption{}
    \label{}
  \end{subfigure}
  \begin{subfigure}{0.11\textwidth}
    \includegraphics[width=\textwidth]{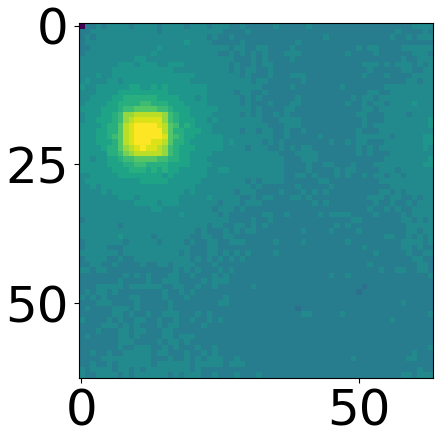}
    \caption{}
    \label{}
  \end{subfigure}
  \begin{subfigure}{0.25\textwidth}
    \includegraphics[width=\textwidth]{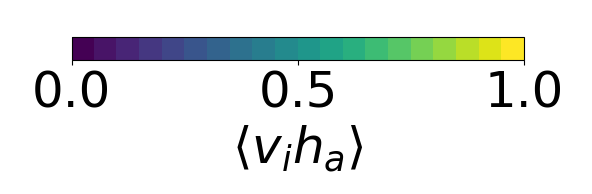}
  \end{subfigure}
  \caption{RG $\corr{vh}$ correlation plot. \\
  Plots (a), (b), (c) and (d) show typical $\corr{v^{(1)}h^{(1)}}$ plots where $v^{(1)}$ is an input spin and $h^{(1)}$ is a block spin produced by one step of RG. \\
  Plots (e), (f), (g) and (h) show typical $\corr{v^{(1)}h^{(2)}}$ plots where ${v}^{(1)}$ is an input spin and $h^{(2)}$ is a  block spin produced by two steps of RG.\\
  Plots (a), (b), (c) and (d) show typical $\corr{v^{(1)}h^{(3)}}$ plots where ${v}^{(1)}$ is an input spin and $h^{(3)}$ is a block spin produced by three steps of RG.\\
In all 12 plots a single block spin's correlator with the complete collection of $64\times 64$ input spins is plotted.}
  \label{fig:rg_corr}
\end{figure}

Figure \ref{fig:rg_corr} shows $\corr{vh}$ plots obtained from the RG flow. Plots (a) to (d) show typical correlation between the input spins and a block spin after one application of RG. The first four plots show a small highly correlated local patch with other areas having lower correlation values. The highly correlated patch corresponds to the set of spins averaged to produce the block spin, so that the coarse graining is indeed reflected in this correlator plot.
Plots (e) to (h) show $\corr{vh}$ plots between the original spins and a block spin produced by two steps of RG. 
The plots clearly demonstrate that the patch of high correlation has increased in size, which is a consequence of the fact that after two steps of RG each block spin averages 16 of the original spins.
Plots (i) to (l) show $\corr{vh}$ plots between the original spins and block spins produced by three steps of RG.
The high correlation region has again increased, reflecting the fact that now each block spin summarizes 64 of the original spins. 
A significant feature of these plots is that the regions of correlation are local, which is a hallmark of the RG coarse graining in which short distance features are removed first by the coarse graining.

\begin{figure}[h!]
  \centering
  \begin{subfigure}{0.11\textwidth}
    \includegraphics[width=\textwidth]{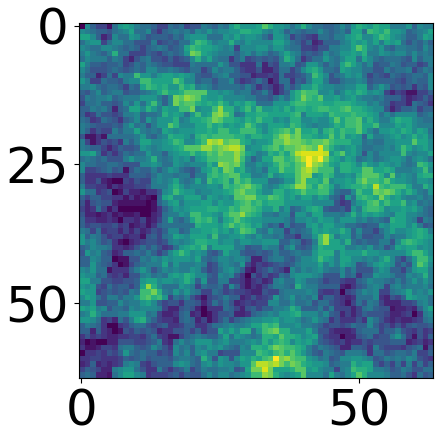}
    \caption{}
    \label{}
  \end{subfigure}
  \begin{subfigure}{0.11\textwidth}
    \includegraphics[width=\textwidth]{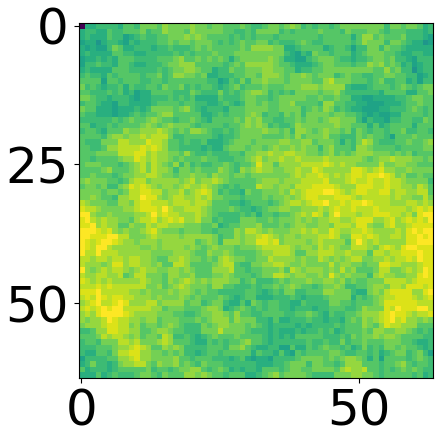}
    \caption{}
    \label{}
  \end{subfigure}
  \begin{subfigure}{0.11\textwidth}
    \includegraphics[width=\textwidth]{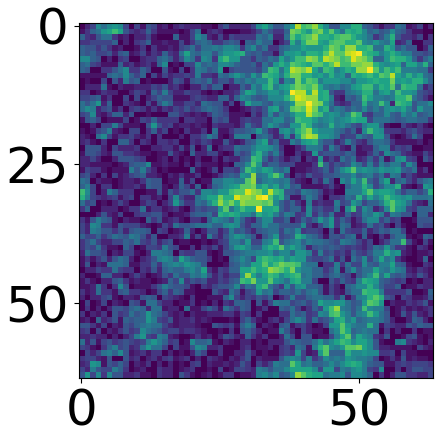}
    \caption{}
    \label{}
  \end{subfigure}
    \begin{subfigure}{0.11\textwidth}
    \includegraphics[width=\textwidth]{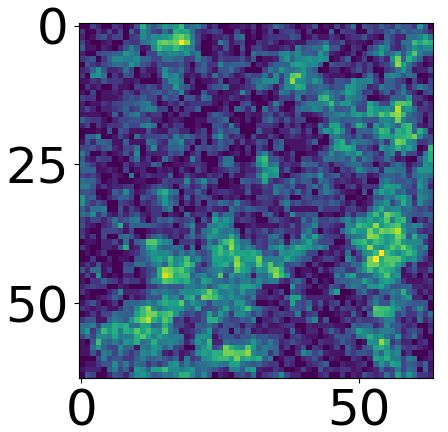}
    \caption{}
    \label{}
  \end{subfigure}

  \begin{subfigure}{0.11\textwidth}
    \includegraphics[width=\textwidth]{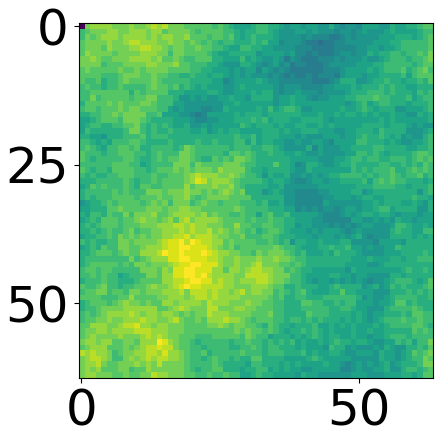}
    \caption{}
    \label{}
  \end{subfigure}
  \begin{subfigure}{0.11\textwidth}
    \includegraphics[width=\textwidth]{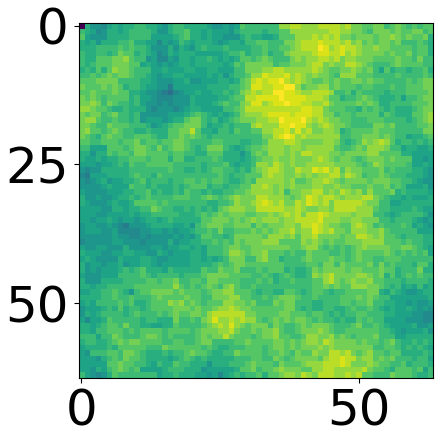}
    \caption{}
    \label{}
  \end{subfigure}
  \begin{subfigure}{0.11\textwidth}
    \includegraphics[width=\textwidth]{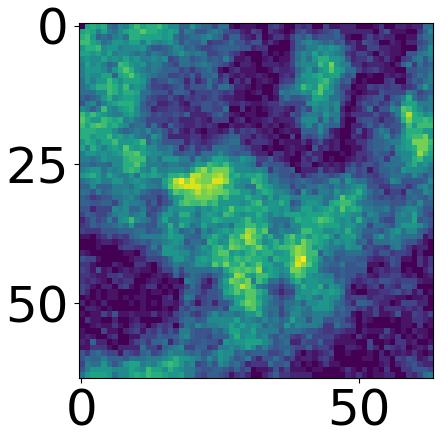}
    \caption{}
    \label{}
  \end{subfigure}
  \begin{subfigure}{0.11\textwidth}
    \includegraphics[width=\textwidth]{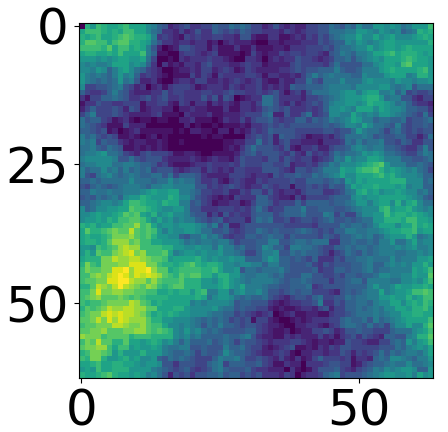}
    \caption{}
    \label{}
  \end{subfigure}

  \begin{subfigure}{0.11\textwidth}
    \includegraphics[width=\textwidth]{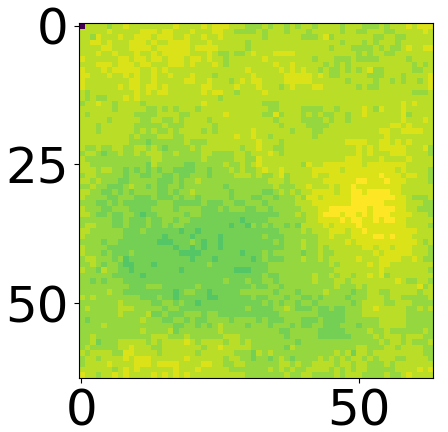}
    \caption{}
    \label{}
  \end{subfigure}
  \begin{subfigure}{0.11\textwidth}
    \includegraphics[width=\textwidth]{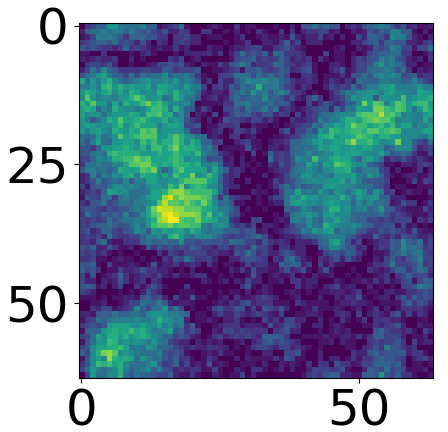}
    \caption{}
    \label{}
  \end{subfigure}
  \begin{subfigure}{0.11\textwidth}
    \includegraphics[width=\textwidth]{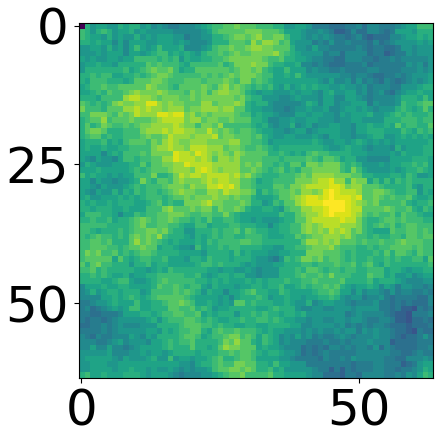}
    \caption{}
    \label{}
  \end{subfigure}
  \begin{subfigure}{0.11\textwidth}
    \includegraphics[width=\textwidth]{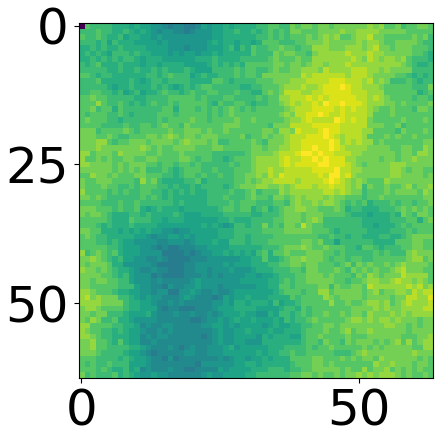}
    \caption{}
    \label{}
  \end{subfigure}
  \begin{subfigure}{0.25\textwidth}
    \includegraphics[width=\textwidth]{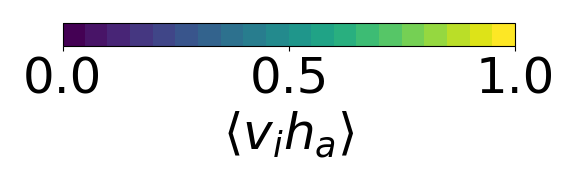}
  \end{subfigure}
  \caption{RBM $\corr{vh}$ correlation plot.\\
  Plots (a), (b), (c) and (d) show a sample of the $\corr{v^{(1)}h^{(1)}}$ plots where $v^{(1)}$ is an input spin and $h^{(1)}$ is a hidden spin produced by the first RBM in the stacked network.\\
  Plots (e), (f), (g) and (h) show a sample of the $\corr{v^{(1)}h^{(2)}}$ plots where $v^{(1)}$ is an input spin and $h^{(2)}$ is a hidden spin produced by the second RBM in the stacked network.\\
    Plots (a), (b), (c) and (d) show a sample of the $\corr{v^{(1)}h^{(3)}}$ plots where $v^{(1)}$ is an input spin and $h^{(3)}$ is a hidden spin produced by the third RBM in the stacked network.\\
    In all 12 plots a single hidden node's correlator with the complete collection of $64\times 64$ input spins is plotted.}
  \label{fig:rbm_corr}
\end{figure}

Figure \ref{fig:rbm_corr} shows $\corr{vh}$ plots for the stacked RBM. 
Again here plots (a) to (d) show $\corr{vh}$ correlation between the training configurations ($64\times64$ visible nodes) and hidden configurations ($32\times32$ hidden nodes) from the first RBM, plots (e) to (h) show $\corr{vh}$ correlations between the training configurations ($64\times64$ visible nodes) and the hidden configurations ($16\times16$ hidden nodes) from the second RBM and plots (i) to (l) show $\corr{vh}$ correlations between configurations ($64\times64$ visible nodes) of the training data and the hidden configurations ($8\times8$ hidden nodes) from the third RBM.
There is no clear difference between the plots for the first layer, second layer and third layer. 
There are some local patches but these are not as distinct as those of the RG flow of Figure \ref{fig:rg_corr}.
Comparison of figures \ref{fig:rg_corr} and \ref{fig:rbm_corr} reveals that there is not much similarity, suggesting that the two coarse graining schemes are quite different.

\begin{figure}[h!]
  \centering
  \begin{subfigure}{0.43\textwidth}
    \includegraphics[width=\textwidth]{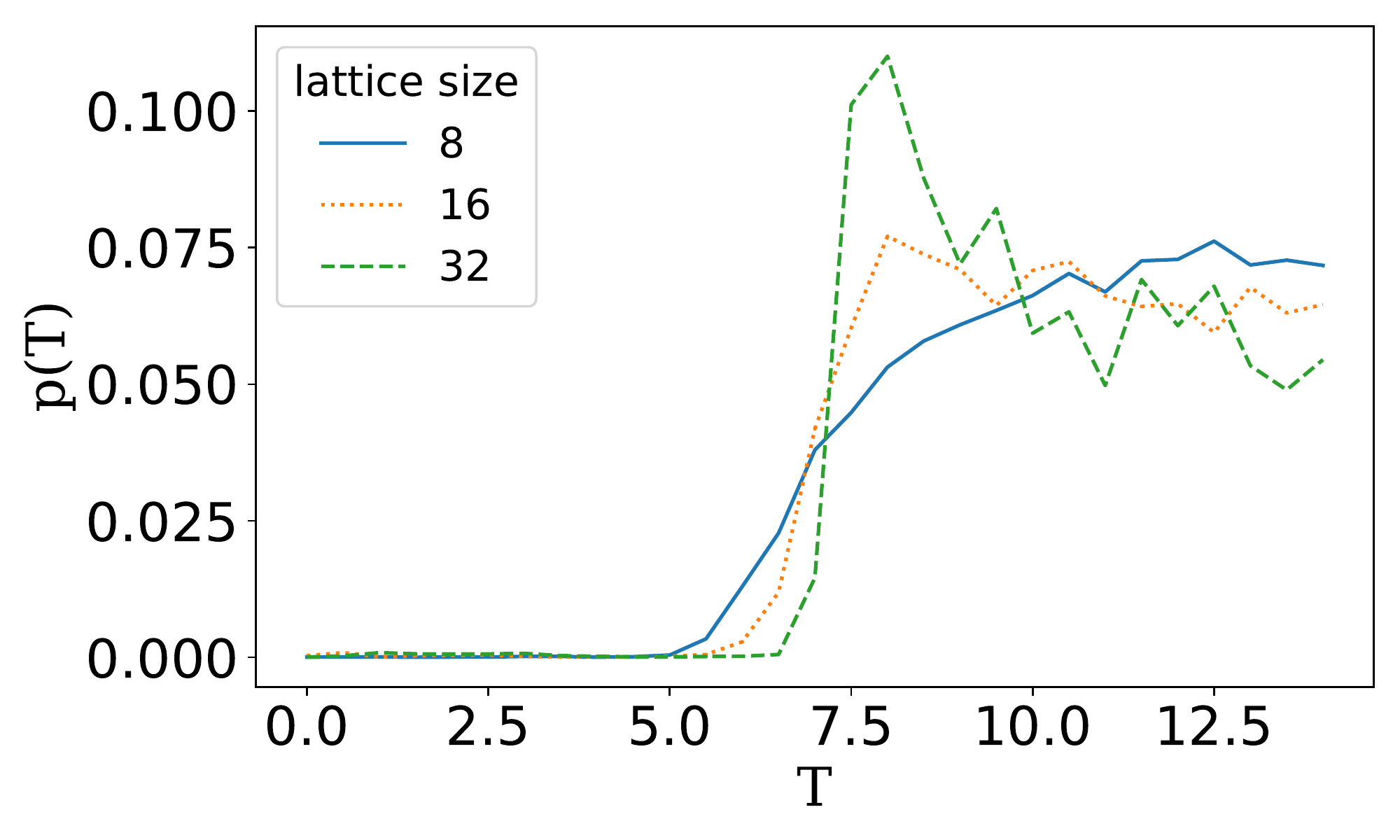}
    \caption{Stacked RBM flow}
    \label{fig:rbm_temp}
  \end{subfigure}
  \begin{subfigure}{0.43\textwidth}
    \includegraphics[width=\textwidth]{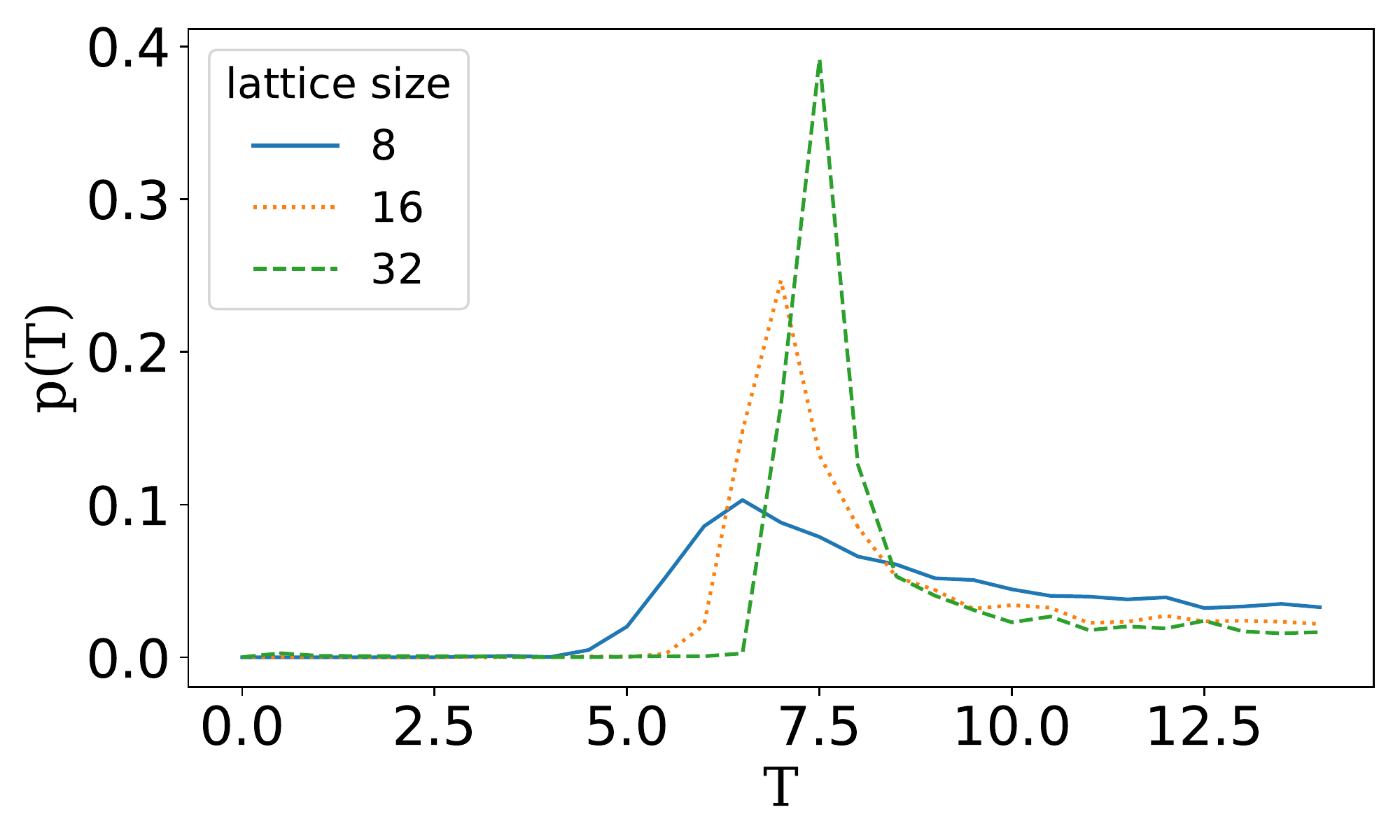}
    \caption{RG flow}
    \label{fig:rg_temp}
  \end{subfigure}
  \caption{Average probability of temperature versus temperature of a flow for a stacked RBM network trained on lattices at $T=7.7$ and an RG flow of 3 steps.
  The RG flow and RBM flows are generated by applying both methods to the training set of $30000$ lattices at $T=7.7$ consisting of $64\times64$ spins.}
  \label{fig:rbm_rg_temp}
\end{figure}

Another interesting aspect to explore is the temperature of each successive flow. 
Here we make use of the same supervised network used to measure temperature in Section \ref{sec:RBMflows}.
Figure \ref{fig:rbm_rg_temp} shows the average measured temperature of configurations at different steps in the RG and stacked RBM flows.
After each step of RG, the temperature of the coarse-grained configurations appears to increase. After one application of RG we measure an average temperature of 6.25 which increases to 7 after two applications of RG and further increases to 7.5 after the third step of RG as seen in Figure \ref{fig:rg_temp}.

The stacked RBM appears to cause a decrease in temperature as we move through successive layers.
The hidden vectors from the first RBM are at a high temperature with a maximum probability at a temperature of 12.5, the second and third RBMs have hidden vectors at a temperature just above 7.5.

Considering the comparison between $\corr{vh}$ plots and the temperature measurement of flows for the RG and stacked RBM network, we do not see convincing evidence that suggest the two are equivalent.
In the case of the measured temperatures, we see very different behavior with the temperature flowing in opposite directions. 
With RG we see temperature increasing with increased steps in the flow and with the stacked RBM we see a decrease in temperature with successive hidden vectors produced by the RBM layers.

For the two-dimensional Ising model as explored in \cite{koch2019deep}, the flows went to a higher temperature in both the case of RG and that of the stacked RBM network.
This contrasts with the flow of the long range spin model where we see an increase in temperature when applying successive steps of RG and a decrease in temperature when applying successive RBMs.

\section{\label{sec:discussion}Conclusions and Discussion}

We have explored the link between RG and deep learning using RBM's trained on data taken from states of a long range spin lattice.
This generalizes existing work \cite{iso2018scale,funai2018thermodynamics} exploring the two-dimensional Ising model with nearest neighbor interactions.
Our results show important differences from those of \cite{iso2018scale,funai2018thermodynamics}, which are discussed and interpreted in this section.\\

We have explored the RBM flow generated by the weights of single RBM network, trained using spin configurations of
a two-dimensional long range spin lattice.
Data sets generated at low temperature, at high temperature and near the critical temperature, have been considered.
Our results demonstrate that regardless of the temperature from which the flow begins, the RBM flow fixed point does not resemble the critical point of the long range spin lattice.
The scaling dimensions for the spin and energy operators do not converge to the values we expect at the critical point.
In addition, when measuring the temperature of the flows, they do not converge to the critical temperature.
This contrasts with results obtained for the Ising spin lattice where the $\Delta_s$ correlator is correctly reproduced and the flows appear to converge to the critical temperature \cite{koch2019deep}.

Our study has also compared RG flow to the flow produced by a stacked deep RBM network.
Following proposals put forward in \cite{koch2019deep} this comparison was performed by comparing the $\corr{vh}$ 
correlator resulting from the two flows.
$\corr{vh}$ plots from RG show local correlated patches surrounded by a large uncorrelated region.
As the number of steps in the RG flow increases so does contrast between the patch of high correlation and the surrounding
uncorrelated environment.
This is very clearly a signature of the locality built into the block spinning RG procedure.
Plots for the stacked RBM show random patches of correlation surrounded by dark uncorrelated regions.
The patches lack the regularity of the patterns the emerge in the case of RG, and consequently they are not enhanced by
subsequent layers of the deep RBM.
As we move through successive layers of the RBM flow's $\corr{vh}$ plots there is no noteworthy oragnized change in the size or shape of the correlated patches which were observed for the RG flow.

In addition to the $\corr{vh}$ plots, the average temperature of successive steps in both the RG and the stacked RBM flows were determined.
As expected the RG flow shows a clear increase in temperature with the length of the flow. 
This simply reflects the presence of relevant operators in the theory so that the fixed point is unstable.
The RBM however shows a different behavior: the temperature decreases with the depth of the stacked network.
For nearest neighbor Ising lattices temperature increases both along the RG flow and with the depth of the stacked RBM
network, showing agreement on this very basic question.
Even for this most basic question, the RBM flow and RG flow disagree for the long ranged spin lattice.

Our results clearly demonstrate that the coarse graining performed by the RBM does not correspond to block spin RG of the long range spin lattice.
For the nearest neighbor Ising model there is at least some evidence that deep RBM's are reproducing some features of the RG flow.
Why does the long ranged spin lattice differ so significantly from the nearest neighbor Ising lattice?
In the long range spin lattice, interactions decrease smoothly in strength as the distance between the interacting spins increases.
The net result is that more distant spins are correlated by interactions in the long range spin lattice, as compared to the Ising 
lattice whose interactions switch off abruptly as one moves beyond nearest neighbors.
Consequently, it is harder for the RBM to recognize locality in the emergent patterns.
Without locality, the resulting coarse graining is nothing like RG.

Setting aside this disagreement, there maybe a more fundamental obstacle to comparing the flows. 
The RBM energy function is simple and linear in $v$ and $h$.
Setting the coefficient of the term linear in $v$ corresponds to fixing the single spin expectation value, i.e. the magnetization of the lattice.
There are no further parameters in the RBM energy function that could be used to fix higher order correlators, such as two point or three point spin correlators.
This implies that not all relevant parameters in the theory have been fixed, so that it would be surprising if the RBM and RG flows terminate at the same point in the space of all possible lattice models.
Small differences between the two will be magnified by the flow making the two look very different.
Including higher order terms in the RBM energy function might allow us to use additional properties present in the input data, during training.
This would allow the RBM to correctly learn essential physical characteristics of the long range spin lattice.
Clearly our most credible conclusion is that more work is required to explore the connection that may exist between RG and deep learning!

\section*{Acknowledgements}

We would like to thank Robert de Mello Koch for useful discussions.


\bibliography{main}

\begin{thebibliography}{37}%
\makeatletter
\providecommand \@ifxundefined [1]{%
 \@ifx{#1\undefined}
}%
\providecommand \@ifnum [1]{%
 \ifnum #1\expandafter \@firstoftwo
 \else \expandafter \@secondoftwo
 \fi
}%
\providecommand \@ifx [1]{%
 \ifx #1\expandafter \@firstoftwo
 \else \expandafter \@secondoftwo
 \fi
}%
\providecommand \natexlab [1]{#1}%
\providecommand \enquote  [1]{``#1''}%
\providecommand \bibnamefont  [1]{#1}%
\providecommand \bibfnamefont [1]{#1}%
\providecommand \citenamefont [1]{#1}%
\providecommand \href@noop [0]{\@secondoftwo}%
\providecommand \href [0]{\begingroup \@sanitize@url \@href}%
\providecommand \@href[1]{\@@startlink{#1}\@@href}%
\providecommand \@@href[1]{\endgroup#1\@@endlink}%
\providecommand \@sanitize@url [0]{\catcode `\\12\catcode `\$12\catcode
  `\&12\catcode `\#12\catcode `\^12\catcode `\_12\catcode `\%12\relax}%
\providecommand \@@startlink[1]{}%
\providecommand \@@endlink[0]{}%
\providecommand \url  [0]{\begingroup\@sanitize@url \@url }%
\providecommand \@url [1]{\endgroup\@href {#1}{\urlprefix }}%
\providecommand \urlprefix  [0]{URL }%
\providecommand \Eprint [0]{\href }%
\providecommand \doibase [0]{http://dx.doi.org/}%
\providecommand \selectlanguage [0]{\@gobble}%
\providecommand \bibinfo  [0]{\@secondoftwo}%
\providecommand \bibfield  [0]{\@secondoftwo}%
\providecommand \translation [1]{[#1]}%
\providecommand \BibitemOpen [0]{}%
\providecommand \bibitemStop [0]{}%
\providecommand \bibitemNoStop [0]{.\EOS\space}%
\providecommand \EOS [0]{\spacefactor3000\relax}%
\providecommand \BibitemShut  [1]{\csname bibitem#1\endcsname}%
\let\auto@bib@innerbib\@empty
\bibitem [{\citenamefont {Krizhevsky}\ \emph {et~al.}(2012)\citenamefont
  {Krizhevsky}, \citenamefont {Sutskever},\ and\ \citenamefont
  {Hinton}}]{krizhevsky2012imagenet}%
  \BibitemOpen
  \bibfield  {author} {\bibinfo {author} {\bibfnamefont {Alex}\ \bibnamefont
  {Krizhevsky}}, \bibinfo {author} {\bibfnamefont {Ilya}\ \bibnamefont
  {Sutskever}}, \ and\ \bibinfo {author} {\bibfnamefont {Geoffrey~E}\
  \bibnamefont {Hinton}},\ }\bibfield  {title} {\enquote {\bibinfo {title}
  {Imagenet classification with deep convolutional neural networks},}\ }in\
  \href@noop {} {\emph {\bibinfo {booktitle} {Advances in neural information
  processing systems}}}\ (\bibinfo {year} {2012})\ pp.\ \bibinfo {pages}
  {1097--1105}\BibitemShut {NoStop}%
\bibitem [{\citenamefont {Vinyals}\ \emph {et~al.}(2019)\citenamefont
  {Vinyals}, \citenamefont {Babuschkin}, \citenamefont {Czarnecki},
  \citenamefont {Mathieu}, \citenamefont {Dudzik}, \citenamefont {Chung},
  \citenamefont {Choi}, \citenamefont {Powell}, \citenamefont {Ewalds},
  \citenamefont {Georgiev} \emph {et~al.}}]{vinyals2019grandmaster}%
  \BibitemOpen
  \bibfield  {author} {\bibinfo {author} {\bibfnamefont {Oriol}\ \bibnamefont
  {Vinyals}}, \bibinfo {author} {\bibfnamefont {Igor}\ \bibnamefont
  {Babuschkin}}, \bibinfo {author} {\bibfnamefont {Wojciech~M}\ \bibnamefont
  {Czarnecki}}, \bibinfo {author} {\bibfnamefont {Micha\"el}\ \bibnamefont
  {Mathieu}}, \bibinfo {author} {\bibfnamefont {Andrew}\ \bibnamefont
  {Dudzik}}, \bibinfo {author} {\bibfnamefont {Junyoung}\ \bibnamefont
  {Chung}}, \bibinfo {author} {\bibfnamefont {David~H}\ \bibnamefont {Choi}},
  \bibinfo {author} {\bibfnamefont {Richard}\ \bibnamefont {Powell}}, \bibinfo
  {author} {\bibfnamefont {Timo}\ \bibnamefont {Ewalds}}, \bibinfo {author}
  {\bibfnamefont {Petko}\ \bibnamefont {Georgiev}},  \emph {et~al.},\
  }\bibfield  {title} {\enquote {\bibinfo {title} {Grandmaster level in
  starcraft ii using multi-agent reinforcement learning},}\ }\href@noop {}
  {\bibfield  {journal} {\bibinfo  {journal} {Nature}\ ,\ \bibinfo {pages}
  {1--5}} (\bibinfo {year} {2019})}\BibitemShut {NoStop}%
\bibitem [{\citenamefont {Ranzato}\ \emph {et~al.}(2010)\citenamefont
  {Ranzato}, \citenamefont {Krizhevsky},\ and\ \citenamefont
  {Hinton}}]{ranzato_factored_2010}%
  \BibitemOpen
  \bibfield  {author} {\bibinfo {author} {\bibfnamefont {Marc’Aurelio}\
  \bibnamefont {Ranzato}}, \bibinfo {author} {\bibfnamefont {Alex}\
  \bibnamefont {Krizhevsky}}, \ and\ \bibinfo {author} {\bibfnamefont
  {Geoffrey}\ \bibnamefont {Hinton}},\ }\bibfield  {title} {\enquote {\bibinfo
  {title} {Factored 3-way restricted boltzmann machines for modeling natural
  images},}\ }in\ \href {http://proceedings.mlr.press/v9/ranzato10a.html}
  {\emph {\bibinfo {booktitle} {Proceedings of the Thirteenth International
  Conference on Artificial Intelligence and Statistics}}}\ (\bibinfo {year}
  {2010})\ pp.\ \bibinfo {pages} {621--628}\BibitemShut {NoStop}%
\bibitem [{\citenamefont {Salakhutdinov}\ \emph {et~al.}(2007)\citenamefont
  {Salakhutdinov}, \citenamefont {Mnih},\ and\ \citenamefont
  {Hinton}}]{salakhutdinov_restricted_2007}%
  \BibitemOpen
  \bibfield  {author} {\bibinfo {author} {\bibfnamefont {Ruslan}\ \bibnamefont
  {Salakhutdinov}}, \bibinfo {author} {\bibfnamefont {Andriy}\ \bibnamefont
  {Mnih}}, \ and\ \bibinfo {author} {\bibfnamefont {Geoffrey}\ \bibnamefont
  {Hinton}},\ }\bibfield  {title} {\enquote {\bibinfo {title} {Restricted
  boltzmann machines for collaborative filtering},}\ }in\ \href {\doibase
  10.1145/1273496.1273596} {\emph {\bibinfo {booktitle} {Proceedings of the
  24th International Conference on Machine Learning}}},\ \bibinfo {series and
  number} {ICML '07}\ (\bibinfo  {publisher} {ACM},\ \bibinfo {address} {New
  York, NY, USA},\ \bibinfo {year} {2007})\ pp.\ \bibinfo {pages} {791--798},\
  \bibinfo {note} {event-place: Corvalis, Oregon, USA}\BibitemShut {NoStop}%
\bibitem [{\citenamefont {Le~Roux}\ and\ \citenamefont
  {Bengio}(2010)}]{le2010deep}%
  \BibitemOpen
  \bibfield  {author} {\bibinfo {author} {\bibfnamefont {Nicolas}\ \bibnamefont
  {Le~Roux}}\ and\ \bibinfo {author} {\bibfnamefont {Yoshua}\ \bibnamefont
  {Bengio}},\ }\bibfield  {title} {\enquote {\bibinfo {title} {Deep belief
  networks are compact universal approximators},}\ }\href@noop {} {\bibfield
  {journal} {\bibinfo  {journal} {Neural computation}\ }\textbf {\bibinfo
  {volume} {22}},\ \bibinfo {pages} {2192--2207} (\bibinfo {year}
  {2010})}\BibitemShut {NoStop}%
\bibitem [{\citenamefont {{Le Roux}}\ and\ \citenamefont
  {Bengio}(2008)}]{le2008representational}%
  \BibitemOpen
  \bibfield  {author} {\bibinfo {author} {\bibfnamefont {Nicolas}\ \bibnamefont
  {{Le Roux}}}\ and\ \bibinfo {author} {\bibfnamefont {Yoshua}\ \bibnamefont
  {Bengio}},\ }\bibfield  {title} {\enquote {\bibinfo {title} {Representational
  power of restricted boltzmann machines and deep belief networks},}\
  }\href@noop {} {\bibfield  {journal} {\bibinfo  {journal} {Neural
  computation}\ }\textbf {\bibinfo {volume} {20}},\ \bibinfo {pages}
  {1631--1649} (\bibinfo {year} {2008})}\BibitemShut {NoStop}%
\bibitem [{\citenamefont {Carrasquilla}\ and\ \citenamefont
  {Melko}(2017)}]{carrasquilla2017machine}%
  \BibitemOpen
  \bibfield  {author} {\bibinfo {author} {\bibfnamefont {Juan}\ \bibnamefont
  {Carrasquilla}}\ and\ \bibinfo {author} {\bibfnamefont {Roger~G}\
  \bibnamefont {Melko}},\ }\bibfield  {title} {\enquote {\bibinfo {title}
  {Machine learning phases of matter},}\ }\href@noop {} {\bibfield  {journal}
  {\bibinfo  {journal} {Nature Physics}\ }\textbf {\bibinfo {volume} {13}},\
  \bibinfo {pages} {431} (\bibinfo {year} {2017})}\BibitemShut {NoStop}%
\bibitem [{\citenamefont {Wang}(2016)}]{wang2016discovering}%
  \BibitemOpen
  \bibfield  {author} {\bibinfo {author} {\bibfnamefont {Lei}\ \bibnamefont
  {Wang}},\ }\bibfield  {title} {\enquote {\bibinfo {title} {Discovering phase
  transitions with unsupervised learning},}\ }\href@noop {} {\bibfield
  {journal} {\bibinfo  {journal} {Physical Review B}\ }\textbf {\bibinfo
  {volume} {94}},\ \bibinfo {pages} {195105} (\bibinfo {year}
  {2016})}\BibitemShut {NoStop}%
\bibitem [{\citenamefont {Deng}\ \emph {et~al.}(2017)\citenamefont {Deng},
  \citenamefont {Li},\ and\ \citenamefont {Sarma}}]{deng2017machine}%
  \BibitemOpen
  \bibfield  {author} {\bibinfo {author} {\bibfnamefont {Dong-Ling}\
  \bibnamefont {Deng}}, \bibinfo {author} {\bibfnamefont {Xiaopeng}\
  \bibnamefont {Li}}, \ and\ \bibinfo {author} {\bibfnamefont {S~Das}\
  \bibnamefont {Sarma}},\ }\bibfield  {title} {\enquote {\bibinfo {title}
  {Machine learning topological states},}\ }\href@noop {} {\bibfield  {journal}
  {\bibinfo  {journal} {Physical Review B}\ }\textbf {\bibinfo {volume} {96}},\
  \bibinfo {pages} {195145} (\bibinfo {year} {2017})}\BibitemShut {NoStop}%
\bibitem [{\citenamefont {Broecker}\ \emph {et~al.}(2017)\citenamefont
  {Broecker}, \citenamefont {Carrasquilla}, \citenamefont {Melko},\ and\
  \citenamefont {Trebst}}]{broecker2017machine}%
  \BibitemOpen
  \bibfield  {author} {\bibinfo {author} {\bibfnamefont {Peter}\ \bibnamefont
  {Broecker}}, \bibinfo {author} {\bibfnamefont {Juan}\ \bibnamefont
  {Carrasquilla}}, \bibinfo {author} {\bibfnamefont {Roger~G}\ \bibnamefont
  {Melko}}, \ and\ \bibinfo {author} {\bibfnamefont {Simon}\ \bibnamefont
  {Trebst}},\ }\bibfield  {title} {\enquote {\bibinfo {title} {Machine learning
  quantum phases of matter beyond the fermion sign problem},}\ }\href@noop {}
  {\bibfield  {journal} {\bibinfo  {journal} {Scientific reports}\ }\textbf
  {\bibinfo {volume} {7}},\ \bibinfo {pages} {8823} (\bibinfo {year}
  {2017})}\BibitemShut {NoStop}%
\bibitem [{\citenamefont {Aoki}\ and\ \citenamefont
  {Kobayashi}(2016)}]{aoki2016restricted}%
  \BibitemOpen
  \bibfield  {author} {\bibinfo {author} {\bibfnamefont {Ken-Ichi}\
  \bibnamefont {Aoki}}\ and\ \bibinfo {author} {\bibfnamefont {Tamao}\
  \bibnamefont {Kobayashi}},\ }\bibfield  {title} {\enquote {\bibinfo {title}
  {Restricted boltzmann machines for the long range ising models},}\
  }\href@noop {} {\bibfield  {journal} {\bibinfo  {journal} {Modern Physics
  Letters B}\ }\textbf {\bibinfo {volume} {30}},\ \bibinfo {pages} {1650401}
  (\bibinfo {year} {2016})}\BibitemShut {NoStop}%
\bibitem [{\citenamefont {Nomura}\ \emph {et~al.}(2017)\citenamefont {Nomura},
  \citenamefont {Darmawan}, \citenamefont {Yamaji},\ and\ \citenamefont
  {Imada}}]{nomura2017restricted}%
  \BibitemOpen
  \bibfield  {author} {\bibinfo {author} {\bibfnamefont {Yusuke}\ \bibnamefont
  {Nomura}}, \bibinfo {author} {\bibfnamefont {Andrew~S}\ \bibnamefont
  {Darmawan}}, \bibinfo {author} {\bibfnamefont {Youhei}\ \bibnamefont
  {Yamaji}}, \ and\ \bibinfo {author} {\bibfnamefont {Masatoshi}\ \bibnamefont
  {Imada}},\ }\bibfield  {title} {\enquote {\bibinfo {title} {Restricted
  boltzmann machine learning for solving strongly correlated quantum
  systems},}\ }\href@noop {} {\bibfield  {journal} {\bibinfo  {journal}
  {Physical Review B}\ }\textbf {\bibinfo {volume} {96}},\ \bibinfo {pages}
  {205152} (\bibinfo {year} {2017})}\BibitemShut {NoStop}%
\bibitem [{\citenamefont {Morningstar}\ and\ \citenamefont
  {Melko}(2017)}]{morningstar2017deep}%
  \BibitemOpen
  \bibfield  {author} {\bibinfo {author} {\bibfnamefont {Alan}\ \bibnamefont
  {Morningstar}}\ and\ \bibinfo {author} {\bibfnamefont {Roger~G}\ \bibnamefont
  {Melko}},\ }\bibfield  {title} {\enquote {\bibinfo {title} {Deep learning the
  ising model near criticality},}\ }\href@noop {} {\bibfield  {journal}
  {\bibinfo  {journal} {The Journal of Machine Learning Research}\ }\textbf
  {\bibinfo {volume} {18}},\ \bibinfo {pages} {5975--5991} (\bibinfo {year}
  {2017})}\BibitemShut {NoStop}%
\bibitem [{\citenamefont {Howard}(2018)}]{howard2018holographic}%
  \BibitemOpen
  \bibfield  {author} {\bibinfo {author} {\bibfnamefont {Eric}\ \bibnamefont
  {Howard}},\ }\bibfield  {title} {\enquote {\bibinfo {title} {Holographic
  renormalization with machine learning},}\ }\href@noop {} {\bibfield
  {journal} {\bibinfo  {journal} {arXiv preprint arXiv:1803.11056}\ } (\bibinfo
  {year} {2018})}\BibitemShut {NoStop}%
\bibitem [{\citenamefont {Paul}\ and\ \citenamefont
  {Venkatasubramanian}(2014)}]{paul_why_2014}%
  \BibitemOpen
  \bibfield  {author} {\bibinfo {author} {\bibfnamefont {Arnab}\ \bibnamefont
  {Paul}}\ and\ \bibinfo {author} {\bibfnamefont {Suresh}\ \bibnamefont
  {Venkatasubramanian}},\ }\bibfield  {title} {\enquote {\bibinfo {title} {Why
  does deep learning work? - a perspective from group theory},}\ }\href
  {http://arxiv.org/abs/1412.6621} {\bibfield  {journal} {\bibinfo  {journal}
  {arXiv:1412.6621 [cs, stat]}\ } (\bibinfo {year} {2014})},\ \bibinfo {note}
  {arXiv: 1412.6621}\BibitemShut {NoStop}%
\bibitem [{\citenamefont {Zhang}\ \emph {et~al.}(2016)\citenamefont {Zhang},
  \citenamefont {Bengio}, \citenamefont {Hardt}, \citenamefont {Recht},\ and\
  \citenamefont {Vinyals}}]{zhang_understanding_2016}%
  \BibitemOpen
  \bibfield  {author} {\bibinfo {author} {\bibfnamefont {Chiyuan}\ \bibnamefont
  {Zhang}}, \bibinfo {author} {\bibfnamefont {Samy}\ \bibnamefont {Bengio}},
  \bibinfo {author} {\bibfnamefont {Moritz}\ \bibnamefont {Hardt}}, \bibinfo
  {author} {\bibfnamefont {Benjamin}\ \bibnamefont {Recht}}, \ and\ \bibinfo
  {author} {\bibfnamefont {Oriol}\ \bibnamefont {Vinyals}},\ }\bibfield
  {title} {\enquote {\bibinfo {title} {Understanding deep learning requires
  rethinking generalization},}\ }\href {http://arxiv.org/abs/1611.03530}
  {\bibfield  {journal} {\bibinfo  {journal} {arXiv:1611.03530 [cs]}\ }
  (\bibinfo {year} {2016})},\ \bibinfo {note} {arXiv: 1611.03530}\BibitemShut
  {NoStop}%
\bibitem [{\citenamefont {Jordan}\ and\ \citenamefont
  {Mitchell}(2015)}]{jordan2015machine}%
  \BibitemOpen
  \bibfield  {author} {\bibinfo {author} {\bibfnamefont {Michael~I}\
  \bibnamefont {Jordan}}\ and\ \bibinfo {author} {\bibfnamefont {Tom~M}\
  \bibnamefont {Mitchell}},\ }\bibfield  {title} {\enquote {\bibinfo {title}
  {Machine learning: Trends, perspectives, and prospects},}\ }\href@noop {}
  {\bibfield  {journal} {\bibinfo  {journal} {Science}\ }\textbf {\bibinfo
  {volume} {349}},\ \bibinfo {pages} {255--260} (\bibinfo {year}
  {2015})}\BibitemShut {NoStop}%
\bibitem [{\citenamefont {Lin}\ \emph {et~al.}(2017)\citenamefont {Lin},
  \citenamefont {Tegmark},\ and\ \citenamefont {Rolnick}}]{lin_why_2017}%
  \BibitemOpen
  \bibfield  {author} {\bibinfo {author} {\bibfnamefont {Henry~W.}\
  \bibnamefont {Lin}}, \bibinfo {author} {\bibfnamefont {Max}\ \bibnamefont
  {Tegmark}}, \ and\ \bibinfo {author} {\bibfnamefont {David}\ \bibnamefont
  {Rolnick}},\ }\bibfield  {title} {\enquote {\bibinfo {title} {Why does deep
  and cheap learning work so well?}}\ }\href {\doibase
  10.1007/s10955-017-1836-5} {\bibfield  {journal} {\bibinfo  {journal}
  {Journal of Statistical Physics}\ }\textbf {\bibinfo {volume} {168}},\
  \bibinfo {pages} {1223--1247} (\bibinfo {year} {2017})},\ \bibinfo {note}
  {arXiv: 1608.08225}\BibitemShut {NoStop}%
\bibitem [{\citenamefont {Mehta}\ and\ \citenamefont
  {Schwab}(2014)}]{mehta_exact_2014}%
  \BibitemOpen
  \bibfield  {author} {\bibinfo {author} {\bibfnamefont {Pankaj}\ \bibnamefont
  {Mehta}}\ and\ \bibinfo {author} {\bibfnamefont {David~J.}\ \bibnamefont
  {Schwab}},\ }\bibfield  {title} {\enquote {\bibinfo {title} {An exact mapping
  between the variational renormalization group and deep learning},}\ }\href
  {http://arxiv.org/abs/1410.3831} {\bibfield  {journal} {\bibinfo  {journal}
  {arXiv:1410.3831 [cond-mat, stat]}\ } (\bibinfo {year} {2014})},\ \bibinfo
  {note} {arXiv: 1410.3831}\BibitemShut {NoStop}%
\bibitem [{\citenamefont {Iso}\ \emph {et~al.}(2018)\citenamefont {Iso},
  \citenamefont {Shiba},\ and\ \citenamefont {Yokoo}}]{iso2018scale}%
  \BibitemOpen
  \bibfield  {author} {\bibinfo {author} {\bibfnamefont {Satoshi}\ \bibnamefont
  {Iso}}, \bibinfo {author} {\bibfnamefont {Shotaro}\ \bibnamefont {Shiba}}, \
  and\ \bibinfo {author} {\bibfnamefont {Sumito}\ \bibnamefont {Yokoo}},\
  }\bibfield  {title} {\enquote {\bibinfo {title} {Scale-invariant feature
  extraction of neural network and renormalization group flow},}\ }\href@noop
  {} {\bibfield  {journal} {\bibinfo  {journal} {Physical Review E}\ }\textbf
  {\bibinfo {volume} {97}},\ \bibinfo {pages} {053304} (\bibinfo {year}
  {2018})}\BibitemShut {NoStop}%
\bibitem [{\citenamefont {Koch}\ \emph {et~al.}(2019)\citenamefont {Koch},
  \citenamefont {Koch},\ and\ \citenamefont {Cheng}}]{koch2019deep}%
  \BibitemOpen
  \bibfield  {author} {\bibinfo {author} {\bibfnamefont {Ellen de~Mello}\
  \bibnamefont {Koch}}, \bibinfo {author} {\bibfnamefont {Robert de~Mello}\
  \bibnamefont {Koch}}, \ and\ \bibinfo {author} {\bibfnamefont {Ling}\
  \bibnamefont {Cheng}},\ }\bibfield  {title} {\enquote {\bibinfo {title} {Is
  deep learning an rg flow?}}\ }\href@noop {} {\bibfield  {journal} {\bibinfo
  {journal} {arXiv preprint arXiv:1906.05212}\ } (\bibinfo {year}
  {2019})}\BibitemShut {NoStop}%
\bibitem [{\citenamefont {Li}\ and\ \citenamefont {Wang}(2018)}]{li2018neural}%
  \BibitemOpen
  \bibfield  {author} {\bibinfo {author} {\bibfnamefont {Shuo-Hui}\
  \bibnamefont {Li}}\ and\ \bibinfo {author} {\bibfnamefont {Lei}\ \bibnamefont
  {Wang}},\ }\bibfield  {title} {\enquote {\bibinfo {title} {Neural network
  renormalization group},}\ }\href@noop {} {\bibfield  {journal} {\bibinfo
  {journal} {Physical review letters}\ }\textbf {\bibinfo {volume} {121}},\
  \bibinfo {pages} {260601} (\bibinfo {year} {2018})}\BibitemShut {NoStop}%
\bibitem [{\citenamefont {Kim}\ and\ \citenamefont
  {Kim}(2018)}]{kim2018smallest}%
  \BibitemOpen
  \bibfield  {author} {\bibinfo {author} {\bibfnamefont {Dongkyu}\ \bibnamefont
  {Kim}}\ and\ \bibinfo {author} {\bibfnamefont {Dong-Hee}\ \bibnamefont
  {Kim}},\ }\bibfield  {title} {\enquote {\bibinfo {title} {Smallest neural
  network to learn the ising criticality},}\ }\href@noop {} {\bibfield
  {journal} {\bibinfo  {journal} {Physical Review E}\ }\textbf {\bibinfo
  {volume} {98}},\ \bibinfo {pages} {022138} (\bibinfo {year}
  {2018})}\BibitemShut {NoStop}%
\bibitem [{\citenamefont {Funai}\ and\ \citenamefont
  {Giataganas}(2018)}]{funai2018thermodynamics}%
  \BibitemOpen
  \bibfield  {author} {\bibinfo {author} {\bibfnamefont {Shotaro~Shiba}\
  \bibnamefont {Funai}}\ and\ \bibinfo {author} {\bibfnamefont {Dimitrios}\
  \bibnamefont {Giataganas}},\ }\bibfield  {title} {\enquote {\bibinfo {title}
  {Thermodynamics and feature extraction by machine learning},}\ }\href@noop {}
  {\bibfield  {journal} {\bibinfo  {journal} {arXiv preprint arXiv:1810.08179}\
  } (\bibinfo {year} {2018})}\BibitemShut {NoStop}%
\bibitem [{\citenamefont {Koch-Janusz}\ and\ \citenamefont
  {Ringel}(2018)}]{koch2018mutual}%
  \BibitemOpen
  \bibfield  {author} {\bibinfo {author} {\bibfnamefont {Maciej}\ \bibnamefont
  {Koch-Janusz}}\ and\ \bibinfo {author} {\bibfnamefont {Zohar}\ \bibnamefont
  {Ringel}},\ }\bibfield  {title} {\enquote {\bibinfo {title} {Mutual
  information, neural networks and the renormalization group},}\ }\href@noop {}
  {\bibfield  {journal} {\bibinfo  {journal} {Nature Physics}\ }\textbf
  {\bibinfo {volume} {14}},\ \bibinfo {pages} {578} (\bibinfo {year}
  {2018})}\BibitemShut {NoStop}%
\bibitem [{\citenamefont {Bény}(2013)}]{beny_deep_2013}%
  \BibitemOpen
  \bibfield  {author} {\bibinfo {author} {\bibfnamefont {Cédric}\ \bibnamefont
  {Bény}},\ }\bibfield  {title} {\enquote {\bibinfo {title} {Deep learning and
  the renormalization group},}\ }\href@noop {} {\bibfield  {journal} {\bibinfo
  {journal} {arXiv preprint arXiv:1301.3124 [quant-ph]}\ } (\bibinfo {year}
  {2013})}\BibitemShut {NoStop}%
\bibitem [{\citenamefont {Saremi}\ and\ \citenamefont
  {Sejnowski}(2013)}]{saremi_hierarchical_2013}%
  \BibitemOpen
  \bibfield  {author} {\bibinfo {author} {\bibfnamefont {Saeed}\ \bibnamefont
  {Saremi}}\ and\ \bibinfo {author} {\bibfnamefont {Terrence~J.}\ \bibnamefont
  {Sejnowski}},\ }\bibfield  {title} {\enquote {\bibinfo {title} {Hierarchical
  model of natural images and the origin of scale invariance},}\ }\href
  {\doibase 10.1073/pnas.1222618110} {\bibfield  {journal} {\bibinfo  {journal}
  {Proceedings of the National Academy of Sciences of the United States of
  America}\ }\textbf {\bibinfo {volume} {110}},\ \bibinfo {pages} {3071--3076}
  (\bibinfo {year} {2013})}\BibitemShut {NoStop}%
\bibitem [{\citenamefont {Geyer}(1992)}]{geyer1992practical}%
  \BibitemOpen
  \bibfield  {author} {\bibinfo {author} {\bibfnamefont {Charles~J}\
  \bibnamefont {Geyer}},\ }\bibfield  {title} {\enquote {\bibinfo {title}
  {Practical markov chain monte carlo},}\ }\href@noop {} {\bibfield  {journal}
  {\bibinfo  {journal} {Statistical science}\ ,\ \bibinfo {pages} {473--483}}
  (\bibinfo {year} {1992})}\BibitemShut {NoStop}%
\bibitem [{\citenamefont {Poland}\ \emph {et~al.}(2019)\citenamefont {Poland},
  \citenamefont {Rychkov},\ and\ \citenamefont {Vichi}}]{poland2019conformal}%
  \BibitemOpen
  \bibfield  {author} {\bibinfo {author} {\bibfnamefont {David}\ \bibnamefont
  {Poland}}, \bibinfo {author} {\bibfnamefont {Slava}\ \bibnamefont {Rychkov}},
  \ and\ \bibinfo {author} {\bibfnamefont {Alessandro}\ \bibnamefont {Vichi}},\
  }\bibfield  {title} {\enquote {\bibinfo {title} {The conformal bootstrap:
  Theory, numerical techniques, and applications},}\ }\href@noop {} {\bibfield
  {journal} {\bibinfo  {journal} {Reviews of Modern Physics}\ }\textbf
  {\bibinfo {volume} {91}},\ \bibinfo {pages} {015002} (\bibinfo {year}
  {2019})}\BibitemShut {NoStop}%
\bibitem [{\citenamefont {Wilson}\ and\ \citenamefont
  {Kogut}(1974)}]{wilson1974renormalization}%
  \BibitemOpen
  \bibfield  {author} {\bibinfo {author} {\bibfnamefont {Kenneth~G}\
  \bibnamefont {Wilson}}\ and\ \bibinfo {author} {\bibfnamefont {John}\
  \bibnamefont {Kogut}},\ }\bibfield  {title} {\enquote {\bibinfo {title} {The
  renormalization group and the $\epsilon$ expansion},}\ }\href@noop {}
  {\bibfield  {journal} {\bibinfo  {journal} {Physics reports}\ }\textbf
  {\bibinfo {volume} {12}},\ \bibinfo {pages} {75--199} (\bibinfo {year}
  {1974})}\BibitemShut {NoStop}%
\bibitem [{\citenamefont {Efrati}\ \emph {et~al.}(2014)\citenamefont {Efrati},
  \citenamefont {Wang}, \citenamefont {Kolan},\ and\ \citenamefont
  {Kadanoff}}]{efrati2014real}%
  \BibitemOpen
  \bibfield  {author} {\bibinfo {author} {\bibfnamefont {Efi}\ \bibnamefont
  {Efrati}}, \bibinfo {author} {\bibfnamefont {Zhe}\ \bibnamefont {Wang}},
  \bibinfo {author} {\bibfnamefont {Amy}\ \bibnamefont {Kolan}}, \ and\
  \bibinfo {author} {\bibfnamefont {Leo~P}\ \bibnamefont {Kadanoff}},\
  }\bibfield  {title} {\enquote {\bibinfo {title} {Real-space renormalization
  in statistical mechanics},}\ }\href@noop {} {\bibfield  {journal} {\bibinfo
  {journal} {Reviews of Modern Physics}\ }\textbf {\bibinfo {volume} {86}},\
  \bibinfo {pages} {647} (\bibinfo {year} {2014})}\BibitemShut {NoStop}%
\bibitem [{\citenamefont {Kadanoff}\ \emph {et~al.}(1976)\citenamefont
  {Kadanoff}, \citenamefont {Houghton},\ and\ \citenamefont
  {Yalabik}}]{kadanoff_variational_1976}%
  \BibitemOpen
  \bibfield  {author} {\bibinfo {author} {\bibfnamefont {Leo~P.}\ \bibnamefont
  {Kadanoff}}, \bibinfo {author} {\bibfnamefont {Anthony}\ \bibnamefont
  {Houghton}}, \ and\ \bibinfo {author} {\bibfnamefont {Mehmet~C.}\
  \bibnamefont {Yalabik}},\ }\bibfield  {title} {\enquote {\bibinfo {title}
  {Variational approximations for renormalization group transformations},}\
  }\href {\doibase 10.1007/BF01011765} {\bibfield  {journal} {\bibinfo
  {journal} {Journal of Statistical Physics}\ }\textbf {\bibinfo {volume}
  {14}},\ \bibinfo {pages} {171--203} (\bibinfo {year} {1976})}\BibitemShut
  {NoStop}%
\bibitem [{\citenamefont {Hinton}(2012)}]{hinton2012practical}%
  \BibitemOpen
  \bibfield  {author} {\bibinfo {author} {\bibfnamefont {Geoffrey~E}\
  \bibnamefont {Hinton}},\ }\bibfield  {title} {\enquote {\bibinfo {title} {A
  practical guide to training restricted boltzmann machines},}\ }in\ \href@noop
  {} {\emph {\bibinfo {booktitle} {Neural networks: Tricks of the trade}}}\
  (\bibinfo  {publisher} {Springer},\ \bibinfo {year} {2012})\ pp.\ \bibinfo
  {pages} {599--619}\BibitemShut {NoStop}%
\bibitem [{\citenamefont {Hinton}(2002)}]{hinton2002training}%
  \BibitemOpen
  \bibfield  {author} {\bibinfo {author} {\bibfnamefont {Geoffrey~E}\
  \bibnamefont {Hinton}},\ }\bibfield  {title} {\enquote {\bibinfo {title}
  {Training products of experts by minimizing contrastive divergence},}\
  }\href@noop {} {\bibfield  {journal} {\bibinfo  {journal} {Neural
  computation}\ }\textbf {\bibinfo {volume} {14}},\ \bibinfo {pages}
  {1771--1800} (\bibinfo {year} {2002})}\BibitemShut {NoStop}%
\bibitem [{\citenamefont {Sutskever}\ and\ \citenamefont
  {Tieleman}(2010)}]{sutskever2010convergence}%
  \BibitemOpen
  \bibfield  {author} {\bibinfo {author} {\bibfnamefont {Ilya}\ \bibnamefont
  {Sutskever}}\ and\ \bibinfo {author} {\bibfnamefont {Tijmen}\ \bibnamefont
  {Tieleman}},\ }\bibfield  {title} {\enquote {\bibinfo {title} {On the
  convergence properties of contrastive divergence},}\ }in\ \href@noop {}
  {\emph {\bibinfo {booktitle} {Proceedings of the thirteenth international
  conference on artificial intelligence and statistics}}}\ (\bibinfo {year}
  {2010})\ pp.\ \bibinfo {pages} {789--795}\BibitemShut {NoStop}%
\bibitem [{\citenamefont {Carreira-Perpinan}\ and\ \citenamefont
  {Hinton}(2005)}]{carreira2005contrastive}%
  \BibitemOpen
  \bibfield  {author} {\bibinfo {author} {\bibfnamefont {Miguel~A}\
  \bibnamefont {Carreira-Perpinan}}\ and\ \bibinfo {author} {\bibfnamefont
  {Geoffrey~E}\ \bibnamefont {Hinton}},\ }\bibfield  {title} {\enquote
  {\bibinfo {title} {On contrastive divergence learning.}}\ }in\ \href@noop {}
  {\emph {\bibinfo {booktitle} {Aistats}}},\ Vol.~\bibinfo {volume} {10}\
  (\bibinfo {organization} {Citeseer},\ \bibinfo {year} {2005})\ pp.\ \bibinfo
  {pages} {33--40}\BibitemShut {NoStop}%
\bibitem [{\citenamefont {Bengio}\ \emph {et~al.}(2006)\citenamefont {Bengio},
  \citenamefont {Lamblin}, \citenamefont {Popovici},\ and\ \citenamefont
  {Larochelle}}]{bengio_greedy_2006}%
  \BibitemOpen
  \bibfield  {author} {\bibinfo {author} {\bibfnamefont {Yoshua}\ \bibnamefont
  {Bengio}}, \bibinfo {author} {\bibfnamefont {Pascal}\ \bibnamefont
  {Lamblin}}, \bibinfo {author} {\bibfnamefont {Dan}\ \bibnamefont {Popovici}},
  \ and\ \bibinfo {author} {\bibfnamefont {Hugo}\ \bibnamefont {Larochelle}},\
  }\bibfield  {title} {\enquote {\bibinfo {title} {Greedy layer-wise training
  of deep networks},}\ }in\ \href
  {http://dl.acm.org/citation.cfm?id=2976456.2976476} {\emph {\bibinfo
  {booktitle} {Proceedings of the 19th {International} {Conference} on {Neural}
  {Information} {Processing} {Systems}}}},\ \bibinfo {series and number}
  {{NIPS}'06}\ (\bibinfo  {publisher} {MIT Press},\ \bibinfo {address}
  {Cambridge, MA, USA},\ \bibinfo {year} {2006})\ pp.\ \bibinfo {pages}
  {153--160},\ \bibinfo {note} {event-place: Canada}\BibitemShut {NoStop}%
\end{thebibliography}%

\end{document}